\documentclass[10pt,twocolumn,letterpaper]{article}
\usepackage{cuted}
\usepackage{CJKutf8}
\usepackage{float}
\usepackage[final]{cvpr}   
\usepackage{xcolor}  
\definecolor{iccvblue}{rgb}{0.21,0.49,0.74}
\usepackage[pagebackref,breaklinks,colorlinks,linkcolor=iccvblue,citecolor=iccvblue,urlcolor=iccvblue]{hyperref}
\usepackage{cuted}
\usepackage{amssymb}
\usepackage{amsmath}
\usepackage{booktabs}
\usepackage{graphicx}
\usepackage{tabularx} 
\usepackage{multirow}
\usepackage{wasysym}
\usepackage{pifont}
\usepackage{adjustbox}

\def\confName{CVPR}
\def\confYear{2025}

\title{FonTS: Text Rendering with Typography and Style Controls}

\author{
    Wenda Shi\textsuperscript{1} \quad
    Yiren Song\textsuperscript{2} \quad
    Dengming Zhang\textsuperscript{3} \quad
    Jiaming Liu\textsuperscript{4} \quad
    Xingxing Zou\textsuperscript{1 $^*$} \vspace{0.5em} \\
    \textsuperscript{1}The Hong Kong Polytechnic University \quad 
    \textsuperscript{2}National University of Singapore \\ 
    \textsuperscript{3}Zhejiang University \quad 
    \textsuperscript{4}Tiamat AI \quad 
    \textsuperscript{$^*$}Corresponding author \\
}

\begin{document}

\maketitle
\begin{strip}
\centering
    \vspace{-16mm}
    \includegraphics[width=\linewidth]{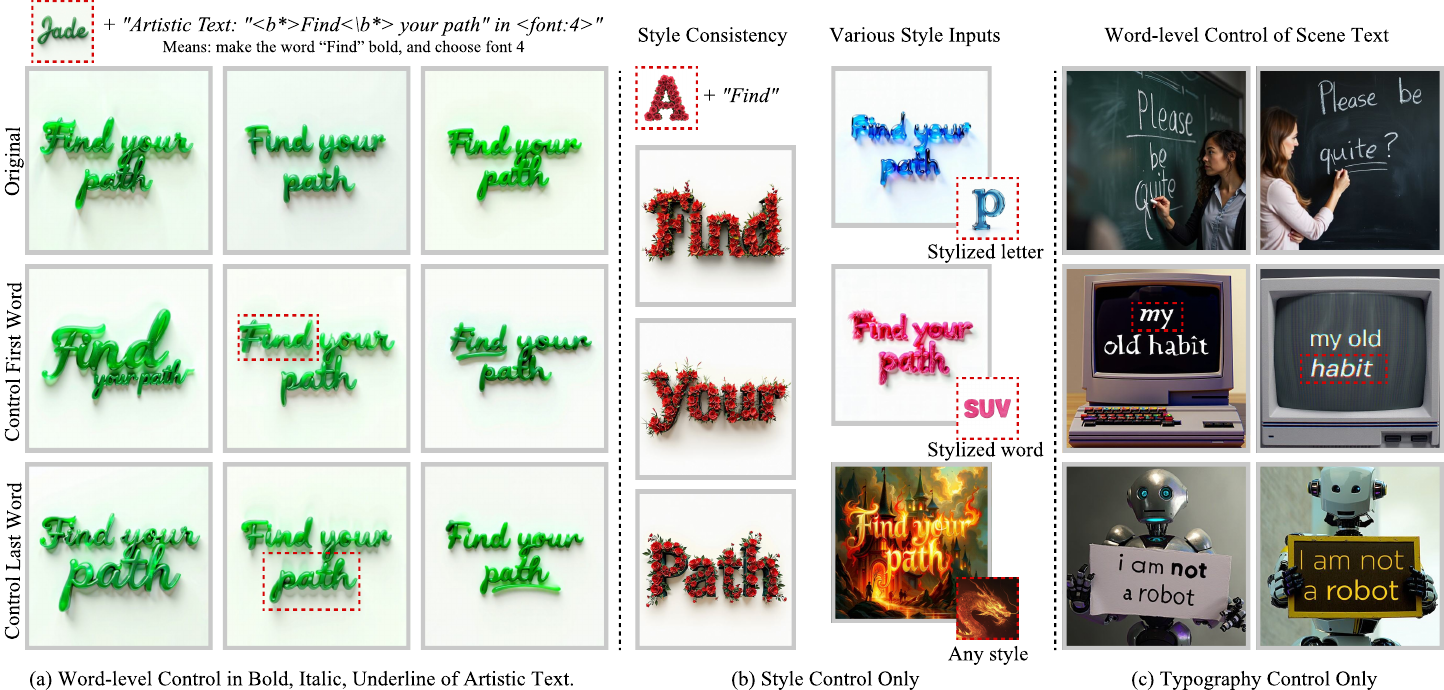}
    \vspace{-7mm}
    \captionof{figure}{Text rendering with typography and style controls. The desired style is indicated by an image, and the prompt defines the text content, including font and word-level attributes. The modifier token—\textless b*\textgreater and \textless \textbackslash b*\textgreater for bold, \textless i*\textgreater and \textless \textbackslash i*\textgreater for italic, \textless u*\textgreater and \textless \textbackslash u*\textgreater for underline—enclosed word to denote the application of effects. Results show that our method effectively supports (a) word-level control and style control, (b) style control only, (c) word-level control without compromising the performance of scene text rendering.}
    \label{fig:teaser}
\end{strip}

\begin{abstract}
Visual text rendering is widespread in various real-world applications, requiring careful font selection and typographic choices. Recent progress in diffusion transformer (DiT)-based text-to-image (T2I) models shows promise in automating these processes. 
However, these methods still encounter challenges like inconsistent fonts, style variation, and limited fine-grained control, particularly at the word-level. 
This paper proposes a two-stage DiT-based pipeline to address these problems by enhancing controllability over typography and style in text rendering. 
We introduce typography control fine-tuning (TC-FT), a parameter-efficient fine-tuning method (on $5\%$ key parameters) with enclosing typography control tokens (ETC-tokens), which enables precise word-level application of typographic features.
To further address style inconsistency in text rendering, we propose a text-agnostic style control adapter (SCA) that prevents content leakage while enhancing style consistency.
To implement TC-FT effectively, we incorporated an HTML-rendered data pipeline and proposed the first word-level controllable dataset.
Through comprehensive experiments, we demonstrate the effectiveness of our approach in achieving superior word-level typographic control, font consistency, and style consistency in text rendering tasks. Our project page is available at \href{https://wendashi.github.io/FonTS-Page/}{\textcolor{iccvblue}{this site.}}
\end{abstract} 
\section{Introduction}
\label{sec:intro}

Visual text images are ubiquitous in daily life and hold significant commercial value in advertising, branding, and marketing~\cite{bai2024intelligent, chen2024textdiffuser}.
However, the design process for visual text is complex and time-consuming. Designers must carefully select appropriate fonts, use typographic elements like italics, and create artistic styles that are aesthetically pleasing and coherent.
Recent advances in diffusion models~\cite{rombach2022high, podell2024sdxl} demonstrate promising potential for creating visual content in design, thereby attracting substantial attention. Concurrently, real-world applications raise increasing demands for control over the generated content~\cite{liao2024appearance, shi2025generative, zhu2025any}.

Previous efforts have mainly focused on improving control over the content accuracy of scene text rendering~\cite{yang2024glyphcontrol, tuo2024anytext, chen2023textdiffuser, chen2024textdiffuser}. 
With the development of DiT-based T2I models, e.g., SD3~\cite{esser2024scaling} and Flux.1~\cite{blackforestlabs2024}, the accuracy of text content has seen significant improvements. 
Beyond content accuracy, Glyph-ByT5~\cite{liu2024glyph} introduced a new text encoder through contrastive learning~\cite{zhuo2023whitenedcse}, enabling various font types of text. 
Textdiffuser-2~\cite{chen2024textdiffuser} trained both two language models and the whole diffusion model to acquire layout planning capabilities. 
While these methods~\cite{liu2024glyph, chen2024textdiffuser} have implemented control at the paragraph-level, no methods have yet realized word-level control.
Moreover, prior methods often overlook the artistic aspects of text ~\cite{chen2024textdiffuser}. 
Recent DiT models~\cite{esser2024scaling,blackforestlabs2024} have demonstrated promising capabilities in artistic text rendering, yet they face challenges like semantic confusion and style inconsistency.

To expand the boundaries of existing methods (summarized in Table \ref{tab:rw-1}), this paper identifies three essential requirements of text rendering methods: 1) control of fonts and word-level attributes in Basic Text Rendering (BTR); 2) consistency in style control in Artistic Text Rendering (ATR); 3) preservation of Scene Text Rendering (STR) capabilities without negative impact.

To this end, we propose a two-stage DiT-based pipeline for text rendering with typography and style controls. For typography control, we introduce Typography Control (TC)-finetuning, a parameter-efficient fine-tuning method, alongside enclosing typography control tokens (ETC-tokens). By introducing HTML-render to ingeniously design the data synthesis pipeline, we propose the first word-level typography control dataset (TC-dataset). Our findings show that the model not only learns typographic elements but also applies specific typographic features at precise word locations. For style control, we introduce a style control adapter (SCA) that injects style information without compromising the accuracy of the text. The training of SCA is also a two-stage process, each stage using a different dataset. In total, these datasets consist of approximately 600k image-text pairs with high aesthetic scores.

We validate the effectiveness of the proposed methods. First, we demonstrate that the learned ETC-tokens can generate text images with the desired word-level typographic attributes, through GPT-4o and manual verification. Next, we assess font consistency in BTR and style consistency in ATR by user studies and quantitative metrics. These evaluations show that our method outperforms various baselines in terms of font consistency and word-level controllability for BTR, and style consistency for ATR.

In summary, our contributions are as follows:

\begin{itemize}
\item We are the first to address the challenge of word-level control in text rendering, via introducing a two-stage DiT-based pipeline that ensures consistency in font and style while preserving scene text rendering capabilities. 
\item We propose a parameter-efficient fine-tuning technique that enables DiT-based T2I models to achieve precise control over local visual details, such as word-level typographic attributes. To address style inconsistency, we design a text-agnostic SCA that prevents content leakage while enhancing style consistency. 
\item We introduce the first word-level controllable dataset. By leveraging ETC-tokens, we enable precise learning of typographic attributes and their specific locations.
\item Our approach outperforms existing baselines, demonstrating superior performance in text rendering while achieving enhanced control over typography and style.
\end{itemize}

\begin{figure*}[htbp]
\centering
    \includegraphics[width=0.8\linewidth]{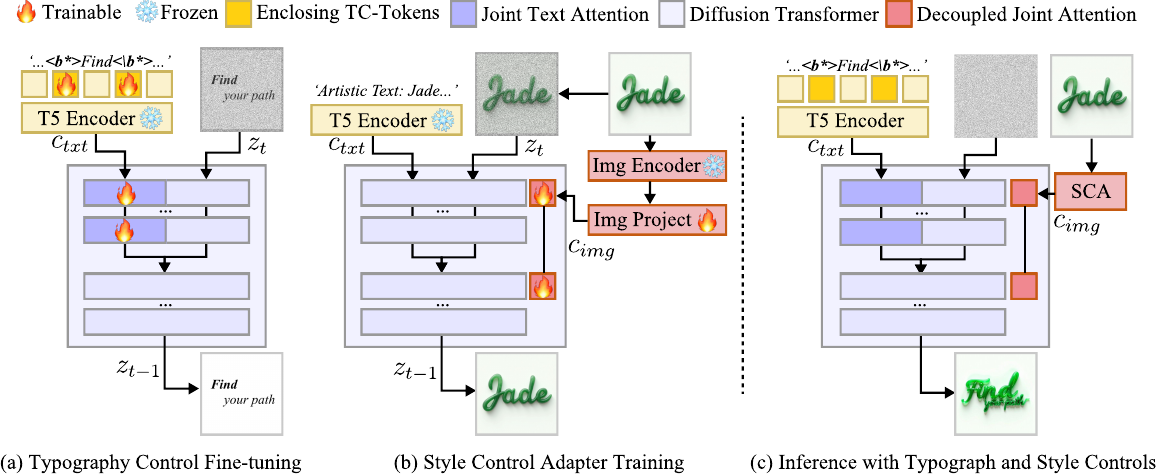}
    \vspace{-3mm}
    \caption{Framework Overview. In the training phase, (a) illustrates the typography control (TC)-finetuning with paired TC-datasets, and (b) presents the training process for style control adapters (SCA). For inference, (c) shows the integrated operation of the TC-finetuned backbone and the SCA. For simplicity, CLIP is omitted in the figure. The prompt in (a) is  `\textless b*\textgreater Find\textless\textbackslash b*\textgreater your path in Font: \textless font:3\textgreater.', and the prompt in (b) is `Artistic Text: 'Jade', the letters are composed of jade, 3d render, minimalist, high resolution, typography'.}
    \vspace{-5mm}
    \label{fig:method-1}
\end{figure*}

\section{Related Work}
\label{sec:related_work}


\begin{table}[t]
\begin{tabular}{m{4cm}m{0.8cm}m{0.8cm}m{0.8cm}}
\toprule
Methods \textbackslash Tasks & BTR & STR & ATR \\
\midrule
Ds-Fusion {\color{gray}\footnotesize[ICCV 23]} & \ding{55} & \ding{55} & \ding{51} \\
Font-Studio {\color{gray}\footnotesize[ECCV 24]} & \ding{55} & \ding{55} & \ding{51} \\
AnyText {\color{gray}\footnotesize[ICLR 24]} & \ding{51} & \ding{51} & \ding{55} \\
Textdiffusers-2 {\color{gray}\footnotesize[ECCV 24]} & \ding{51} & \ding{51} & \ding{55} \\
Glyph-ByT5 {\color{gray}\footnotesize[ECCV 24]} & \ding{51} & \ding{51} & \ding{55} \\
SD3 / Flux {\color{gray}\footnotesize[ICML 24]} & \ding{51} & \ding{51} & \ding{51} \\
Ours & \ding{51}+C & \ding{51}+C & \ding{51}+C \\
\bottomrule
\end{tabular}
\vspace{-2mm}
\caption{Differences with existing methods, C means controls.}
\label{tab:rw-1}
\vspace{-5mm}
\end{table}

\noindent \textbf{Scene Text Rendering.} Despite progress in diffusion models \cite{rombach2022high, podell2024sdxl}, high-quality scene text rendering remains a challenge. To address this, prior research~\cite{yang2024glyphcontrol, chen2024textdiffuser, chen2023textdiffuser, anytext24} focuses on explicitly controlling the position and content of the text being rendered, relying on ControlNet~\cite{zhang2023adding}. 
Another line of works~\cite{liu2024glyph, liu2024glyph2} fine-tunes the character-aware ByT5 text encoder \cite{liu-etal-2023-character} using paired glyph-text datasets, improving the ability to render accurate text in images.

\noindent \textbf{Artistic Text Rendering.} Early research focused on font creation by transferring textures from existing characters, employing stroke-based methods~\cite{berio2022strokestyles, song2022clipfont, song2023clipvg}, patch-based techniques~\cite{yang2017awesome, yang2018context, yang2018context2}, and GAN-based~\cite{azadi2018multi, jiang2019scfont, gao2019artistic, Yang_2019_ICCV, Mao2023IntelligentTA, Tang_2022_CVPR, wu2019editing} methods.  
Innovations with diffusion models~\cite{tanveer2023ds, wang2023anything, mu2024fontstudio, shi2025wordcon, lu2025easytext} have enabled diverse text image stylization and semantic typography, resulting in visually appealing designs that retain readability. However, despite recent DiT models \cite{esser2024scaling, blackforestlabs2024} showing quite promise in artistic text rendering, they struggle with semantic confusion and style inconsistency.

\noindent \textbf{Controllable Image Generation.} Controllable generation methods~\cite{song2024processpainter, zhang2024ssr, zhang2025stable, zhang2024stable, wang2024stablegarment, chen2025transanimate, ma2024followyouremoji, ma2025followyourclick, ma2024followpose, ma2025followcreation, ma2025followyourmotion, zhuo2024vividdreamer, zhuo2024infinidreamer} enable control over diffusion models to synthesize specific subjects or layouts, typically by fine-tuning on user-provided examples and modifying the attention mechanism. For text-based controllable methods~\cite{gal2023an, ruiz2023dreambooth, kumari2023multi}, ColorPeel~\cite{butt2025colorpeel} constructs color-shape pairs to generate images with target colors. 
For image-based controllable methods, UniControl~\cite{qin2023unicontrol} retrain T2I models from scratch, which is computationally expensive~\cite{zhang2023adding}. A more efficient alternative involves integrating trainable modules into existing models as adapters, enabling structural ~\cite{ye2023ip, mou2024t2i, liao2024attentional} and style controls~\cite{chen2024artadapter, gong2025relationadapter, zhou2024stylefactory, liao2024uni}. 

Most prior approaches are implemented on U-Net with a single CLIP text encoder. While existing methods have explored controllable generation under the Diffusion Transformer (DiT) architecture ~\cite{zhang2025easycontrol, tan2024ominicontrol, song2025layertracer, song2025makeanything, huang2025photodoodle, song2025omniconsistency, guo2025any2anytryon, wang2025diffdecompose, wan2024grid}, such studies remain relatively limited. Moreover, the area of controllable generation under multi-modal conditions has not been well-explored. Our work advances this direction by enabling word-level controls and seamlessly integrating multi-modal controls to broaden applications.

\section{Approach}
Our proposed pipeline trains distinct components for different objectives to achieve uniquely balance between the content accuracy and stylization. The proposed parameter-efficient fine-tuning method with enclosing typography control tokens (ETC-tokens), shown in Figure~\ref{fig:method-1} (a), provides word-level controls under resource constraints. Meanwhile, style control adapters training (in Figure~\ref{fig:method-1}(b)) overcomes the content leakage in style control.

\subsection{Typography Control Learning}
\noindent \textbf{Preliminaries of Rectified Flow DiT.}
To avoid the computationally expensive process of ordinary differential equation (ODE), diffusion transformers such as \cite{esser2024scaling, blackforestlabs2024} directly regress a vector field $u_t$ that generates a probability path between noise distribution $p_1$ and data distribution $p_0$. To construct such a vector field $u_t$, \cite{esser2024scaling} consider a forward process that corresponds to a probability path $p_t$ transitioning from $p_0$ to $p_1=\mathcal{N}(0, 1)$. This can be represented as $z_t = a_t x_0 + b_t \epsilon\quad\text{, where}\;\epsilon \sim \mathcal{N}(0,I)$.
With the conditions $a_0 = 1, b_0 = 0, a_1 = 0$ and $b_1 = 1$, the marginals $p_t(z_t) =
  \mathbb{E}_{\epsilon \sim \mathcal{N}(0,I)}
  p_t(z_t \vert \epsilon)\;$ align with data and noise distribution. Referring to \cite{lipman2023flow,esser2024scaling}, the marginal vector field $u_t$ can generate the marginal probability paths $p_t$, using the conditional vector fields as follows:
\begin{align}
    u_t(z) = \mathbb{E}_{\epsilon \sim
  \mathcal{N}(0,I)} u_t(z \vert \epsilon) \frac{p_t(z \vert
  \epsilon)}{p_t(z)},
  \label{eq:marginal_u}
\end{align}
The conditional flow matching objective is formulated as:
\begin{align}
   {L_{CFM}} =  \mathbb{E}_{t, p_t(z | \epsilon), p(\epsilon) }|| v_{\Theta}(z, t) - u_t(z | \epsilon)  ||_2^2\;, \label{eq:condflowmatch}
\end{align}
where the conditional vector fields $u_t(z \vert \epsilon)$ provides a tractable and equivalent objective. 

\begin{figure}[t]
  \centering
   \includegraphics[width=\linewidth]{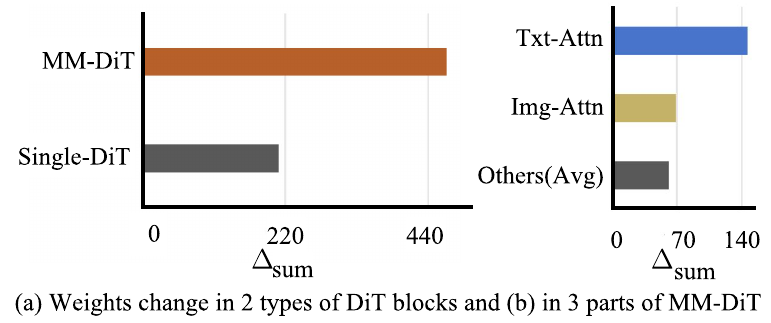}
   \vspace{-7mm}
   \caption{Comparative weight changes in the transformer backbone during full parameter fine-tuning. (a) shows that the MM-DiT experiences double the weight changes compared to the Single-DiT. (b) indicates that the Txt-Attn also shows the double weight changes relative to other components within the MM-DiT.}
   \vspace{-6mm}
   \label{fig:fig-3}
\end{figure}

\noindent \textbf{Typography Control Fine-tuning.} Previous studies have shown that fine-tuning certain U-Net components can generate specific objects and colors through learned prompts (modifier tokens) within single CLIP text encoder ~\cite{butt2025colorpeel, kumari2023multi, gal2023an}. However, these methods are not applicable to our pipeline. The reason for this lies in the the architectural disparities between DiT and U-Net, and also due to differences between T5 and CLIP. Following \cite{kumari2023multi, li2020few}, we analyzed parameter changes in the fine-tuned transformer backbone on the target dataset for 100k steps using the loss ${L_{CFM}}$ in Eq. \ref{eq:condflowmatch}. The change in parameters for layer $l$ is calculated as $\Delta_l = ||\theta_{l}' - \theta_{l}||/||\theta_{l}||,$, where $\theta_{l}'$ and $\theta_{l}$ are the fine-tuned and pretrained model parameters, respectively. The total change across all layers is: $\Delta_{sum} = \sum_{l=0}^{n} mean(\Delta_l)$. These parameters are derived from two types of layers: (1) MM-DiT blocks (merging text and image embeddings), and (2) Single-DiT blocks (processing merged embeddings from MM-DiT). In MM-DiT blocks, parameters are divided into three components: joint text attention (Txt-Attn), joint image attention (Img-Attn), and additional modules like multi-layer perception (MLP) and modulation blocks. Figure \ref{fig:fig-3} shows that MM-DiTs have approximately double the weight change of Single-DiTs, with the Txt-Attn component showing nearly twice the change of other MM-DiT elements, despite it is only 5\% parameters of total backbone.

\begin{figure}[t]
  \centering
   \includegraphics[width=\linewidth]{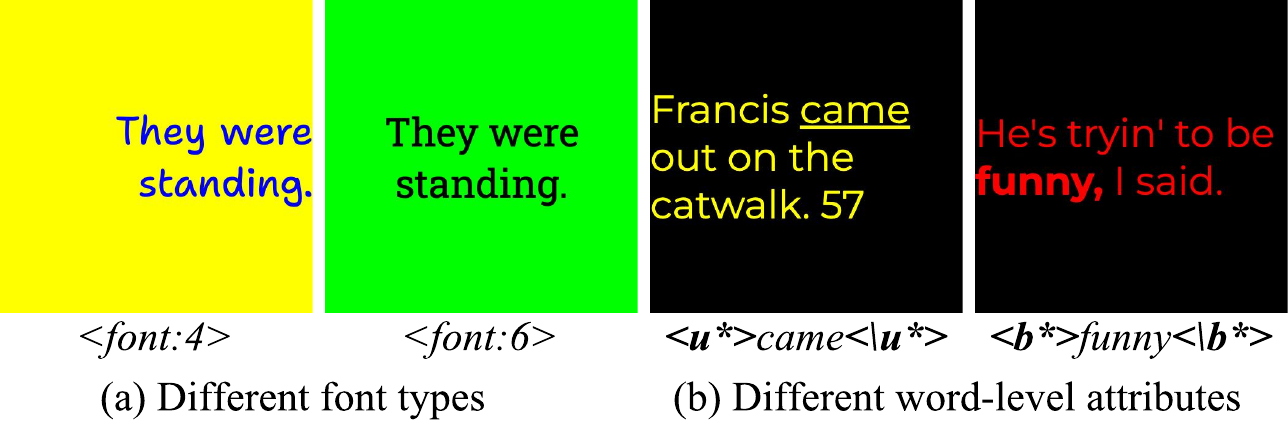}
   \vspace{-7mm}
    \caption{Examples of TC-Dataset featuring two types of TC-Tokens. (a) illustrates the TC-token for various font types. (b) displays the ETC-token with word-level typographic attributes applied to a specific word, including bold, italic, and underline.}
    \vspace{-5mm}
    \label{fig:method-3}
\end{figure}

\noindent \textbf{Enclosing Typography Control (ETC)-Tokens.} We introduce novel modifier tokens for word-level control, to render text with specific typographic feature on targeted words. Our approach differs from previous methods~\cite{kumari2023multi, gal2023an, butt2025colorpeel} in three key ways. 1) Previous methods typically rely on single CLIP text encoder. In contrast, there are two text encoders (CLIP and T5) in our pipeline. This makes design more intricate. We opted to add new modifier tokens only to T5. The reason is that the text embedding from T5 directly feeds into the attention of DiT backbone. In comparison, the text embedding from CLIP only serves as a coarse-graine (pooled) condition, as noted in \cite{esser2024scaling}. 2) T5 and CLIP have distinct characteristics. CLIP has a highly unified space that can align images with text \cite{gal2023an, cho2023promptstyler}, which is not available in T5 trained only on text modality \cite{raffel2020exploring}. Consequently, simply training new modifier tokens for T5 is insufficient. It is essential to carry out cooperative training with other modules. Our ablation study presented in Table \ref{tab:ablation-1} further validates this point. 3) Existing methods use a single modifier token to represent an object~\cite{kumari2023multi, gal2023an} or a type of color~\cite{butt2025colorpeel}, we employ enclosing modifier tokens, each contains a starting token and an ending token, to represent one typographic feature. It allows the model to learn the attribute and its precise application location-a specific word. As shown in Figure \ref{fig:method-3}(b), the enclosing typography control tokens (ETC-tokens) in the example `\textless u*\textgreater came\textless\textbackslash u*\textgreater' indicate an underline effect on the word ``came", localizing the effect to that word alone. These modifier tokens are optimized with joint text attention during fine-tuning.

To fill the gap in high-quality datasets that combine text with typographic attributes, we created the TC-Dataset using typography control rendering (TC-Render). The pipeline leverages HTML rendering to produce images featuring typographic attributes like fonts and word-level styles such as bold, italic, and underline, as shown in Figure \ref{fig:method-3}. Details of the dataset are available in the supplementary Sec {\color{iccvblue}5}.		

\subsection{Style Control Adapters}\label{SCA}
\noindent \textbf{Decoupled Joint Attention.}
The joint attention here refers to the attention in MM-DiT blocks of SD3 \cite{esser2024scaling} and Flux \cite{blackforestlabs2024}. Given the text features $c_{txt}$ and input of joint attention $z_t$, the output of joint attention ${z}'$ can be defined as:
\vspace{-2mm}
\begin{equation}
\vspace{-1.5mm}
\begin{split}
{z}'=\text{Attention}({Q},{K},{V}) = \text{Softmax}(\frac{{Q}{K}^{\top}}{\sqrt{d}}){V},
\\
\end{split}
\end{equation}
where ${Q}=z_c{W}_q$, ${K}=z_c{W}_k$, ${V}=z_c{W}_v$ are the query, key, and values matrices of the attention operation respectively, $z_c = concat({z_t},{c_{txt}})$,and ${W}_q$, ${W}_k$, ${W}_v$ are the weight matrices of the trainable layers.

In order to better decouple style and content, we additionally introduce a decoupled joint attention mechanism (DJA). Inspired by \cite{ye2023ip, chen2024artadapter, mou2024t2i}, we add DJA at the joint attention layers for text features $c_{txt}$ and image features $c_{img}$ are separate. To be specific, we add new joint attention layers in the original MM-DiT and Single-DiT blocks to insert image features. Given the image features $c_{img}$, the output of new joint attention ${z}''$ is as follows:
\vspace{-2mm}
\begin{equation}
    \begin{split}
    {z}''=\text{Attention}({Q}',{K}',{V}') = \text{Softmax}(\frac{{Q'}{K'}^{\top}}{\sqrt{d}}){V}',\\
    \end{split}
\end{equation}
where, ${Q}'={z_t}{W}_q$, ${K}'={c}_{img}{W}'_k$ and ${V}'={c}_{img}{W}'_v$ are the query, key, and values matrices from the image features. ${W}'_k$ and ${W}'_v$ are the corresponding weight matrices. Consequently, we only need to add two parameters ${W}'_k$, ${W}'_v$ for each decoupled joint attention layer.
Then, we simply add the output of image cross-attention to the output of text cross-attention. Hence, the final formulation of the decoupled cross-attention is defined as follows:

\begin{equation}
\begin{split}
{z}^{new}=\text{Softmax}(\frac{{Q}{K}^{\top}}{\sqrt{d}}){V}+ \lambda * \text{Softmax}(\frac{{Q'}{K'}^{\top}}{\sqrt{d}}){V}',\\
\end{split}
\end{equation}
\text{where} ${Q}={z_c}{W}_q, {K}={z_c}{W}_k, {V}={z_c}{W}_v,{Q}'={z_t}{W}_q,
 {K}'={c}_{img}{W}'_k, {V}'={c}_{img}{W}'_v$
, and $\lambda$ represents scale of $c_{img}$. And only ${W}'_k$ and ${W}'_v$ are trainable. 

\noindent \textbf{Style Control Training.}
Style control training consists of two phases, each utilizing different carefully prepared datasets. The phase 1 involves common pretraining with general image-text pairs. Besides, we have introduced phase 2 to better adapt to ATR tasks and avoid content leakage caused by using artistic text images as input. The phase 2 is further fine-tuning after phase 1, using a dataset that includes artistic text images and paired descriptions. For phase 1, we assembled a dataset called \textit{SC-general}, which includes approximately 580k general image-text pairs with high aesthetic scores. These images were sourced from open-source datasets \cite{laionaesthetics, sun2024journeydb}. For phase 2, we created the \textit{SC-artext} dataset. We compile a list of style descriptions and a list of words. Combining these lists generated various prompts for artistic text images, which were then used as input for Flux.1-dev \cite{bai2024intelligent}, resulting in approximately 20k high-quality images. To ensure the images matched the original text content, we used shareGPT4v \cite{chen2023sharegpt4v} to regenerate captions. The datasets are detailed in supplementary Sec {\color{iccvblue} 5}.

\noindent \textbf{Design Choice of Image Encoder.}
In artistic text rendering, borrowing the style of artistic text images is crucial \cite{yang2017awesome, yang2018context, yang2018context2}. But these images carry text information, risking content leakage. To avoid this, the image encoder should be as text-agnostic as possible. Therefore, we select CLIP~\cite{radford2021learning} among alternatives like DINO~\cite{oquab2023dinov2}, Resampler~\cite{awadalla2023openflamingo} or SigLIP~\cite{zhai2023sigmoid} widely used in existing adapers~\cite{chen2024anydoor, ye2023ip}, due to CLIP's visual embeddings are text-insensitive \cite{liu2024glyph, chen2024anydoor}. More discussion are in supplementary Sec {\color{iccvblue} 1.2-(3)}.

\section{Experiments}
\noindent \textbf{Text Rendering Benchmark.}
To assess the text rendering capabilities with word-level typography and style controls, we extend the existing scene text rendering benchmark \cite{chen2023textdiffuser} by introducing new benchmarks for basic text and artistic text rendering.
\noindent \textit{BTR-bench.}
To evaluate word-level typography controls in basic text rendering, we introduce basic text rendering benchmark (BTR-bench). BTR-bench includes 100 prompts of different fonts and typographic attributes. For each text prompt, typographic attributes are randomly applied to three positions within the text to assess the model's ability to render specific typographic attributes on individual words, while font attributes are applied to the entire text in the image.
\noindent \textit{ATR-bench.}
To evaluate artistic text rendering, we introduce the Artistic Text Rendering benchmark (ATR-bench). Similar to the single-letter and multi-letter classification in \cite{tanveer2023ds}, we categorize the content into single-word and multi-word groups. Drawing on the style prompts from the GenerativeFont benchmark \cite{mu2024fontstudio}, we generate artistic individual letters and words using Flux \cite{blackforestlabs2024}. These generated artistic letters and words are used for single-word and multi-word text rendering, respectively.

\noindent \textbf{Implementation Details.}
We use Flux.1-dev \cite{blackforestlabs2024} as the base model for its strong text rendering. In TC-FT, we fix the text prompt for CLIP text encoder since the pooled embedding provides only coarse-graine information \cite{esser2024scaling}. 

\noindent \textbf{Training details.}
For \textit{TC-FT}, we fine-tune the base model for 40k steps using the \textit{TC-Dataset}, incorporating a regularization prefix (`sks') in the text prompts. The total batch size is 32. For the \textit{style control adapters}, we train for 100k steps on \textit{SC-general} and 15k steps on \textit{SC-artext} dataset, using total batch size of 64. All training is on 8 $\times$ A100 and 512 resolution, with a learning rate of $1 \times 10^{-5}$.

\begin{figure*}[t]
\centering
    \includegraphics[width=\linewidth]{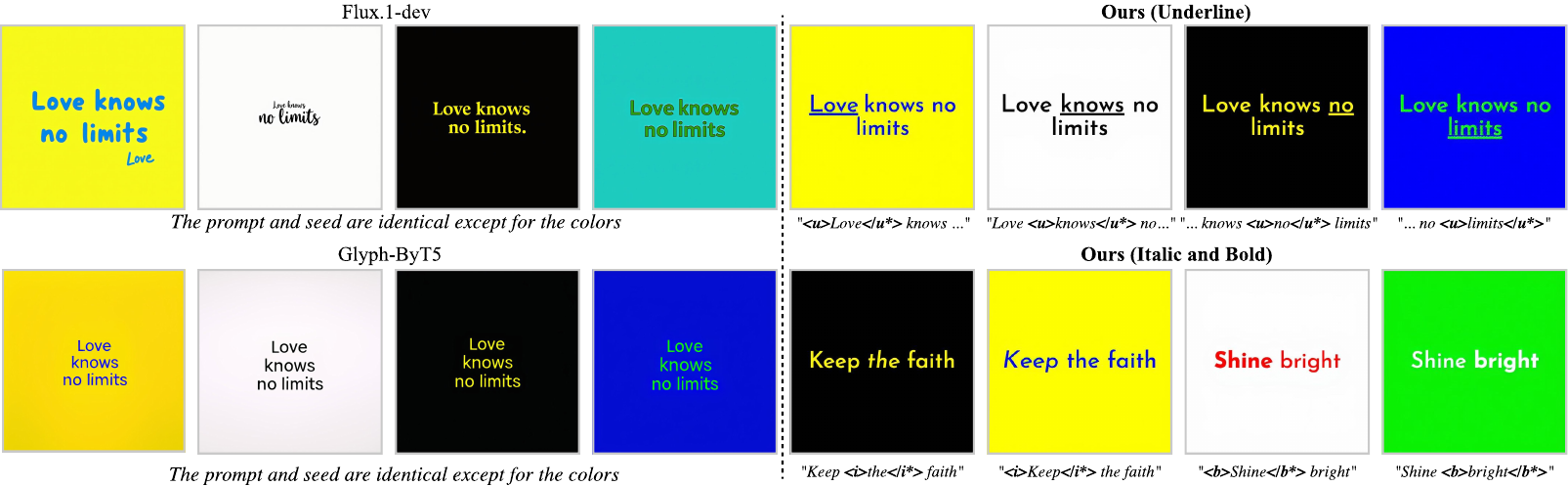}
    \vspace{-6.5mm}
    \caption{Qualitative results on the font consistency and word-level controls in basic text rendering compared with baselines.}
    \vspace{-3mm}
    \label{fig:qual-1}
\end{figure*}

\begin{table}[t]
    \centering
         \begin{tabularx}{\columnwidth}{Xcccc}
            \toprule
            \multirow{2}{*}{Methods} & \multicolumn{2}{c}{Consistency} & \multicolumn{2}{c}{Accuracy} \\
            \cmidrule(lr){2-3} \cmidrule(lr){4-5}
            ~ & \scriptsize FontCLIP-I${\uparrow}$ & \scriptsize Font-Con${\uparrow}$ & \scriptsize Word-Acc${\uparrow}$ & \scriptsize OCR-Acc${\uparrow}$\\
            \midrule
            Glyph \cite{liu2024glyph} & 93.68 & 32.73 & \ding{55} & \textbf{96.36}  \\
            TD-2 \cite{chen2024textdiffuser} & 86.17 & 1.81 & \ding{55} & 42.86  \\
            SD3 \cite{esser2024scaling} & 87.37 & 0.91 & \ding{55} & 48.05  \\
            Flux \cite{blackforestlabs2024} & 90.67 & 0.91 & \ding{55} & 66.49  \\
            \textbf{Ours} & \textbf{96.98} & \textbf{63.64} & \textbf{55.00} & 82.85 \\
            \bottomrule 
        \end{tabularx}
        \vspace{-2mm}
    \caption{Quantitative results for basic text rendering.}
    \vspace{-6mm}
    \label{tab:quant-btr}
\end{table}

\subsection{Quantitative Results}
\noindent \textbf{Quantitative Metrics.} 
We conduct evaluations from two perspectives: consistency and accuracy. For accuracy, we use tool~\cite{PaddleOCR} in \cite{anytext24} to calculate the OCR accuracy (OCR-Acc). In the basic text rendering (BTR), existing OCR tools struggle to evaluate word-level typographic attribute accuracy (Word-Acc). Therefore, we use GPT4o and manual screening to assess and obtain the corresponding score. For consistency, we calculate CLIP image scores (CLIP-I) in artistic text rendering (ATR) and scene text rendering (STR). To better evaluate font consistency, we use FontCLIP \cite{tatsukawa2024fontclip} instead of CLIP, referring to the scores as FontCLIP-I. Beyond automated evaluations, we conduct user studies for font consistency (Font-Con) in BTR and style consistency (Style-Con) in ATR. 

\noindent \textbf{Basic Text Rendering.} 
We compare with Glyph-ByT5 (Glyph) \cite{liu2024glyph}, TextDiffuser-2 (TD-2) \cite{chen2024textdiffuser}, SD3-medium (SD3) \cite{esser2024scaling} and Flux.1-dev (Flux) \cite{blackforestlabs2024} on BTR-bench. As shown in Table \ref{tab:quant-btr}, our method outperformed the baselines in three out of four metrics while slightly below Glyph-ByT5 regarding OCR accuracy in BTR. This is reasonable since Glyph-ByT5 was trained on millions of text images, whereas our approach utilized a dataset of only 50k basic text images, which is twenty times smaller than theirs.

\begin{table}[t]
    \centering
    \begin{tabularx}{\columnwidth}{>{\raggedright\arraybackslash}m{1.7cm}>{\centering\arraybackslash}m{1.25cm}>{\centering\arraybackslash}m{1.85cm}>{\centering\arraybackslash}m{1.85cm}}
        \toprule
        \multirow{2}{*}{Methods} & \multicolumn{2}{c}{Consistency} & \multicolumn{1}{c}{Accuracy} \\
        \cmidrule(lr){2-3} \cmidrule(lr){4-4} 
        ~ & \scriptsize CLIP-I${\uparrow}$ & \scriptsize Style-Con${\uparrow}$ & \scriptsize OCR-Acc${\uparrow}$ \\
        \midrule
        SD3 \cite{esser2024scaling} & 60.24 & 13.64 & 24.16 \\
        Flux \cite{blackforestlabs2024} & 64.06 & 17.42 & 48.71 \\
        SD3-IPA & 59.59 & 0 / 9.09 & 6.25 / 9.18 \\
        Flux-IPA & 57.05 & 3.41 / 2.27 & 19.14 / 28.49 \\
        SD3-IPA\tiny+SC & 62.66 & 20.12 / 22.43 & 5.40 / 14.34 \\
        Flux-IPA\tiny+SC & 63.57 & 23.35 / 25.76 & 13.02 / 30.15 \\
        Flux-Redux & 59.56 & 0 / 10.32 & 0 / 5.83 \\
        \textbf{Ours} & \textbf{64.27} & \textbf{31.82} / \textbf{34.09} & \textbf{61.78} / \textbf{59.66} \\
        \bottomrule
    \end{tabularx}
    \vspace{-2mm}
    \caption{Quantitative comparison of artistic text rendering. Ours, SD3-IPA, and Flux-IPA use scale = 0.9/0.6. Redux considers original / interpolation settings. For CLIP-I, the average is reported. SC: style captions. Text prompts for each method in Figure \ref{fig:qual-3}.}
    \vspace{-3mm}
    \label{tab:quant-atr}
\end{table}

\noindent \textbf{Artistic Text Rendering.} 
Comparing with SD3 \cite{esser2024scaling}, Flux \cite{blackforestlabs2024}, SD3 with IP Adapter (SD3-IPA) \footnote{\href{https://huggingface.co/CiaraRowles/IP-Adapter-Instruct}{SD3-IPA}}, Flux-IPA \footnote{\href{https://huggingface.co/InstantX/FLUX.1-dev-IP-Adapter}{Flux-IPA}}, Flux-Redux \footnote{\href{https://blackforestlabs.ai/flux-1-tools/}{Flux-Redux}}, results are shown in Table \ref{tab:quant-atr}. Moreover, we refer to the ComfyUI Node \footnote{\href{https://github.com/kaibioinfo/ComfyUI_AdvancedRefluxControl}{ComfyUI-AdvancedRefluxControl}} and apply interpolation to balance the image prompt with text prompt in Flux-Redux. Despite using a simpler text input, our method outperforms others across all metrics under various hyperparameter settings.

Additionally, we compare with Flux in STR task on the MARIO-bench \cite{chen2023textdiffuser}. As indicated in Table \ref{tab:quant-str}, our method significantly improves OCR-Acc and CLIP scores. Upon reviewing the image results, we found that Flux exhibits semantic confusion in the STR task (on MARIO-bench \cite{chen2023textdiffuser}), which notably reduces its OCR accuracy.

\begin{table}[t]
  \centering
  \begin{tabular}{m{3cm}m{2cm}m{2cm}}
    \toprule
    Methods & OCR-Acc${\uparrow}$ & CLIP-I${\uparrow}$ \\
    \midrule
    Flux \cite{blackforestlabs2024} & 24.17 & 29.93 \\
    \textbf{Ours} & \textbf{53.57} & \textbf{31.66} \\
    \bottomrule
  \end{tabular}
   \vspace{-2mm}
  \caption{Quantitative results for scene text rendering.}
  \label{tab:quant-str}
  \vspace{-5mm}
\end{table}

\noindent \textbf{User Studies.} We conducted user studies to perceptually evaluate our results against baselines on two key aspects: font consistency (Font-Con) and style consistency (Style-Con). Details are provided in the supplementary Sec {\color{iccvblue} 7} .

\begin{figure*}[htbp]
\centering
    \includegraphics[width=\linewidth]{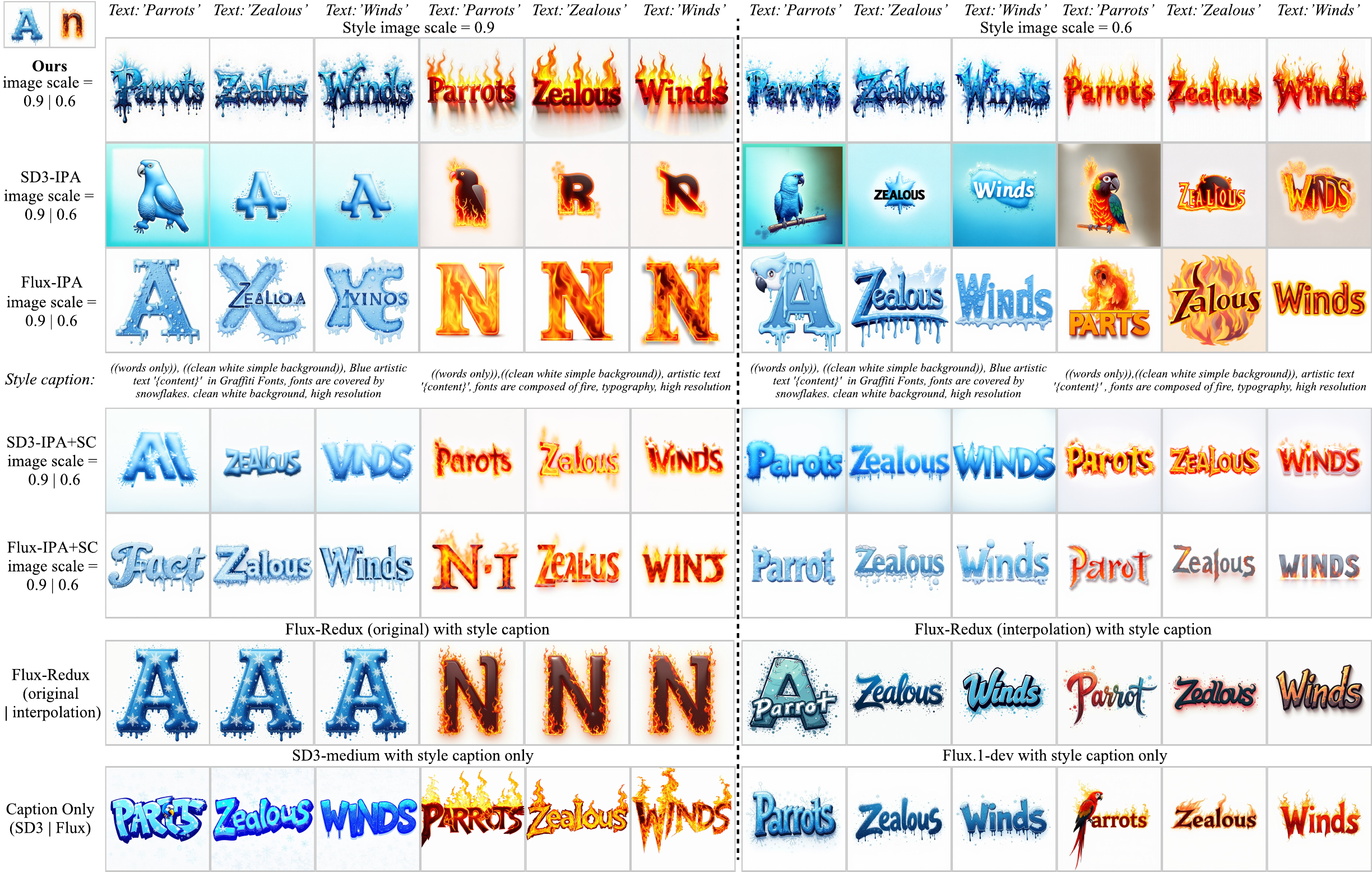}
    \vspace{-3.5mm}
    \caption{Qualitative comparison of style consistency and content accuracy in artistic text rendering against baselines. For all rows except the last row, the input consists of a text prompt along with style images on the top-left. In the top three rows, the text prompts are just simple captions ``Text:`Word'", while for others are style captions.}
    \vspace{-3mm}
    \label{fig:qual-3}
\end{figure*}

\subsection{Qualitative Results}

\noindent \textbf{Basic Text Rendering.} We use a set of challenging prompt words for evaluation. For Flux and Glyph-ByT5, the prompt (for the leftmost images) is: ``Blue Text: `Love knows no limits' in Font: Josefin Sans, Add underline to `Love', Background: pure yellow". This prompt specifies the font and applies a typographic attribute to a word. In Figure \ref{fig:qual-1}, Glyph-ByT5 achieves better font consistency than Flux but lacks word-level control. In contrast, our method ensures strong font consistency and enables word-level control, such as underline, bold, or italic.

\noindent \textbf{Artistic Text Rendering.} To ensure a fair comparison, we set the same seed for each row. The style caption is the same prompt which used to generate the artistic single letters by Flux ( `A' and `n' in top left of Figure \ref{fig:qual-3}). 

Obviously, our results show the best style consistency while preserving the accuracy of the text, comparing with baselines in Table \ref{tab:quant-atr}. In second and third rows of Figure \ref{fig:qual-3}, because the text prompt is relatively simple, output suffers from severe content leakage from style image. There is also semantic confusion, e.g., words `Parrots' becoming parrots itself. In fourth and fifth rows, after using the style caption, the text content becomes prominent. However, the style consistency remains poor, and there are issues with content and capitalization errors, such as `Parots' and `WINDS'). When scale is set to 0.9, content leakage still exists, e.g. the first image in the fourth row  (similar to `A'), and the fourth image in the fifth row (similar to `N'). In Flux-Redux, the results from original is merely about generating variations from style images, and style of results from interpolation is obviously inconsistent. The caption only method lacks style control. As a result, even with the same seed and prompt, the outputs are also inconsistent in style.
In addition, we find that the typography controls in BTR can be transferred to ATR and STR to a certain extent in Figure ~\ref{fig:teaser}. It is reasonable that the degree of controllability will be affected by given style image, particularly when it contains text. However, considering we only use basic text images to learn those word-level attributes, it shows potential for domain generalization ability of proposed method.

\begin{figure}[t]
  \centering
   \includegraphics[width=\linewidth]{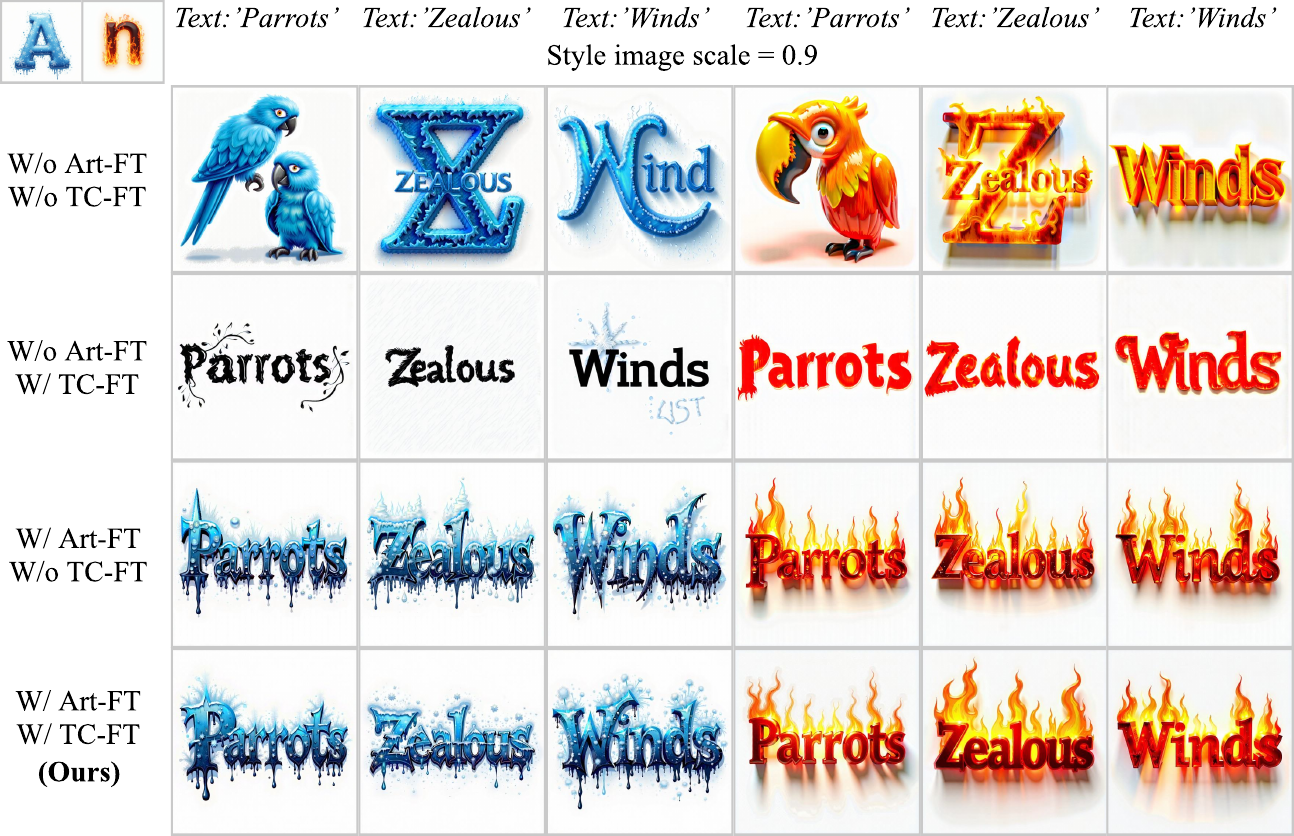}
   \vspace{-3mm}
   \caption{Ablation study of style control adapter (SCA) on second phase finetuing with SC-artext (Art-FT) and TC-FT.}
   \label{fig:ablation_sca_TCFT}
   \vspace{-6mm}
\end{figure}

\subsection{Ablation study}
\noindent \textbf{Ablation on TC-FT.} To assess the effectiveness of typography control fine-tuning (TC-FT), we set up four configurations: training new tokens only, T5 with new tokens, joint text attention (Txt-Attn) with new tokens, and joint text and image attention (Txt+Img-Attn) with new tokens. The results in Table \ref{tab:ablation-1} show the performance in the BTR task. The result of training tokens only is similar to the original Flux. This is likely due to T5 being trained solely on the text modality, lacking the joint vision-language space of CLIP. As stated in \cite{esser2024scaling, liu2024glyph}, text rendering capabilities mainly depend on the text encoder. So we tried to train T5, we found it severely degraded text accuracy (visual results detailed in supplementary Sec {\color{iccvblue} 2.2}). The data in the last two rows indicate that fine-tuning only Txt-Attn is a more effective approach. In training, more parameters typically demand more training steps for better performance. The text rendering capabilities of SD3/Flux also rely on DPO (Direct Preference Optimization) \cite{esser2024scaling}, which we didn't apply. Without DPO, fewer training steps are needed to preserve prior knowledge and mitigate overfitting. As shown in Figure \ref{fig:limi-1}, excessive training steps reduced the in-context nature of text and background in scene text images

\begin{table}[t]
\centering
\scalebox{1}{
\begin{tabularx}{\columnwidth}{Xc c c c c}
\toprule
Trained Modules & \# Para & Ocr-Acc${\uparrow}$  & Word-Acc${\uparrow}$  \\
\midrule
Tokens only & 0.78\% & 66.52 & \ding{55} \\
T5 text encoder & 28.23\% & 0 & \ding{55} \\
Txt-Attn (\textbf{Ours}) & 5.03\% & \textbf{82.85}  & \textbf{55.00} \\
Txt+Img-Attn & 9.29\% & 77.92 & 31.00  \\
\bottomrule 
 \end{tabularx}}
\vspace{-2.5mm}
 \caption{Ablation of different modules during TC-FT on BTR.}
\label{tab:ablation-1}
\vspace{-2.5mm}
\end{table}

\begin{table}[t]
\centering
\scalebox{1}{
\begin{tabularx}{\columnwidth}{c c c c c}
\toprule
Art-FT & TC-FT & CLIP-I ${\uparrow}$ &  OCR-Acc ${\uparrow}$ &  Avg ${\uparrow}$\\
\midrule
\ding{55} & \ding{55} & 60.07 & 28.89 & 44.48 \\
\ding{55} & \ding{51} & 58.09 & \textbf{65.39}  & 61.74 \\
\ding{51} & \ding{55} & \textbf{65.12} & 34.48 & 49.80 \\
\ding{51} & \ding{51} & 64.27 & 60.07 & \textbf{62.17} \\
\bottomrule 
\end{tabularx}}
\vspace{-2.5mm}
\caption{Ablation studies of fine-tuning with SC-artext (Art-FT) for SCA (on MM-DiT and Single-DiT both) and typography control fine-tuning (TC-FT) for backbone. The last row is ours.}
\label{tab:ablation-SCA-both}
\vspace{-2.5mm}
\end{table}

\noindent \textbf{Ablation on SCA.} Style control adapters (SCA) are trained through two phases as mentioned in Sec~\ref{SCA}, and ablation focuses on the second phase. Comparing top two rows in Figure \ref{fig:ablation_sca_TCFT}, it is evident that TC-FT enhances text accuracy, yet severely weakens the artistry. Shifting to the third row, Art-FT significantly boosts artistry without damaging the accuracy. The fourth row, being nearly identical to the third, suggests that after Art-FT, TC-FT has a minimal negative impact on artistry. Results presented in Table~\ref{tab:ablation-SCA-both} further validate this observation in ATR, and we also compared SCA on MM-DiT only, with MM-DiT and Single-DiT both. The results are detailed in supplementary Sec {\color{iccvblue} 2.1}.

\noindent \textbf{Ablation on ETC-Tokens.} The ETC tokens are designed for assigning the words which need to be controlled. We consider three cases: 1) Non-token: directly use the prompt as ``the `word' in Bold''; 2) single token: use a single token; 3) Ours. The results are detailed in supplementary Sec {\color{iccvblue} 2.3}.

\subsection{Applications and Limitation}
\begin{figure}[t]
  \centering
   \includegraphics[width=\linewidth]{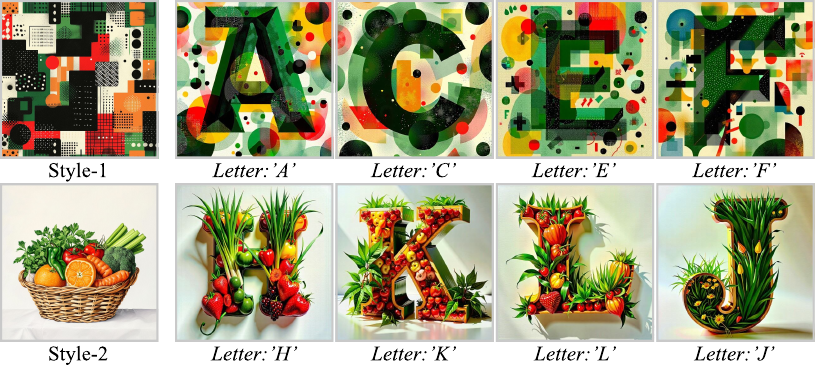}
   \vspace{-6.5mm}
   \caption{The results of artistic letters with different styles.}
   \label{fig:app-0}
   \vspace{-3.5mm}
\end{figure}

\noindent \textbf{Applications.}
\noindent \textit{Artistic font design.}
Benefit from the robust style consistency, our approach can generate a variety of artistic letters with high consistency. Moreover, because the SCA is pre-trained on high-quality, large-scale data, the style control is not limited to artistic text images. Any style image can be used as a control input, as shown in Figure \ref{fig:app-0}.

\begin{figure}[t]
  \centering
   \includegraphics[width=\linewidth]{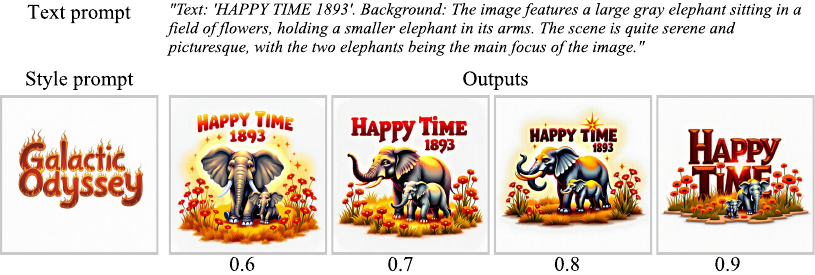}
   \vspace{-3.5mm}
   \caption{The logo design of stylized scene text image with artistic text images and different image scales.}
   \label{fig:app-2}
   \vspace{-3mm}
\end{figure}

\noindent \textit{Logo design.} 
Scene text and artistic text images can also be seamlessly integrated. By using scene text image prompts alongside artistic text images for style control, our method achieves a smooth blend of the two, as shown in Figure~\ref{fig:app-2}. These show our method is suitable for various applications.

\begin{figure}[t]
  \centering
   \includegraphics[width=\linewidth]{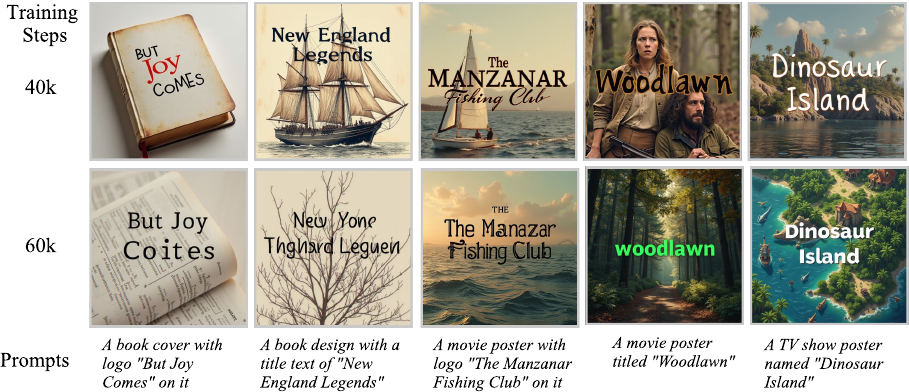}
   \vspace{-3.5mm}
   \caption{Inference results of TC-finetuned models at different training steps. The example prompts are from MARIO-bench~\cite{chen2023textdiffuser}.}
   \vspace{-6mm}
   \label{fig:limi-1}
\end{figure}

\noindent \textbf{Limitation.}
It is observed the language drift phenomenon exists in our method, as the same as \cite{kumari2023multi,ruiz2023dreambooth}. This effect becomes noticeable as the number of training steps increases. This is mainly because, in the TC-FT process, we did not use additional regularization datasets; instead, we applied a simple regularization prefix, `sks', in the text prompts of the TC dataset. This way decreases the cost. As shown in Figure \ref{fig:limi-1}, although language drift is severe at 60k steps, leading to the separation of text and scene in the generated image, the results at 40k steps are acceptable.
\vspace{-2mm}

\section{Conclusion}
We propose a two-stage DiT-based pipeline for text rendering with typography and style controls. TC-FT with ETC-tokens enables the model to learn and apply word-level attributes. The style control adapter facilitates style control without compromising text content. Additionally, we introduce the first word-level control dataset. Experimental results demonstrate that our method outperforms baselines in font consistency and style consistency, and word-level controls for text rendering tasks. This paper is the first to achieve word-level control in text rendering. In future work, we plan to explore its extension to multilingual rendering.

\section*{Acknowledgment}

This work was substantially supported by a grant from the Research Grants Council of the Hong Kong Special Administrative Region, China (Project No. PolyU/RGC Project PolyU 25211424) and partially supported by a grant from PolyU university start-up fund (Project No. P0047675).

{
    \small
    \bibliographystyle{ieeenat_fullname}
    \bibliography{main}
}

\appendix

\maketitlesupplementary
\maketitle



\begin{figure*}[htbp]
\centering
    \includegraphics[width=\linewidth]{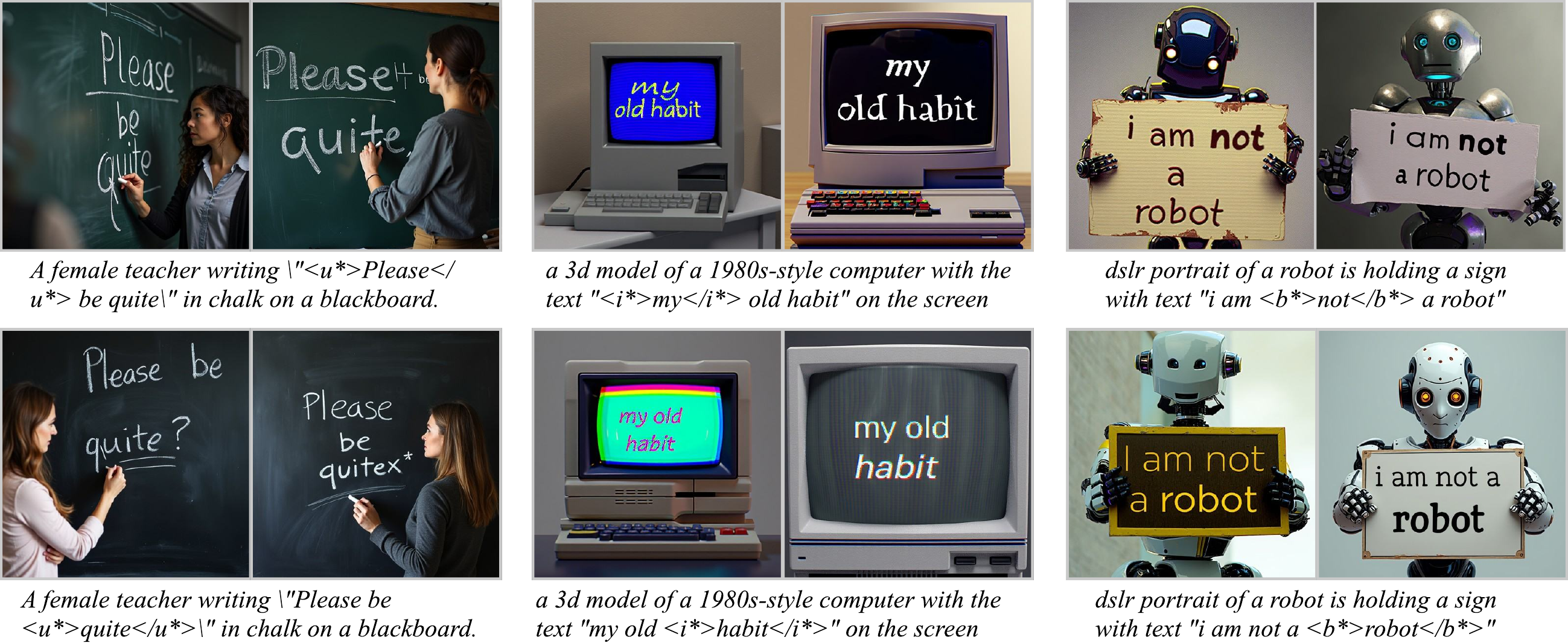}
    \caption{Examples of typographic controls in STR.}
    \label{fig:supp-con-STR}
\end{figure*}

\begin{figure*}[htbp]
\centering
    \includegraphics[width=\linewidth]{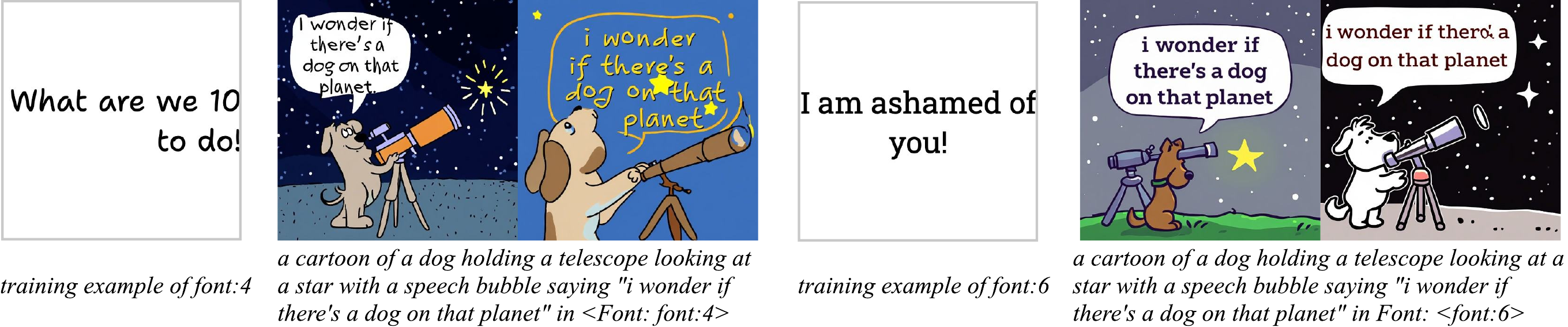}
    \caption{Examples of font selection in STR.}
    \label{fig:supp-font-STR}
\end{figure*}

\begin{figure*}[htbp]
\centering
    \includegraphics[width=\linewidth]{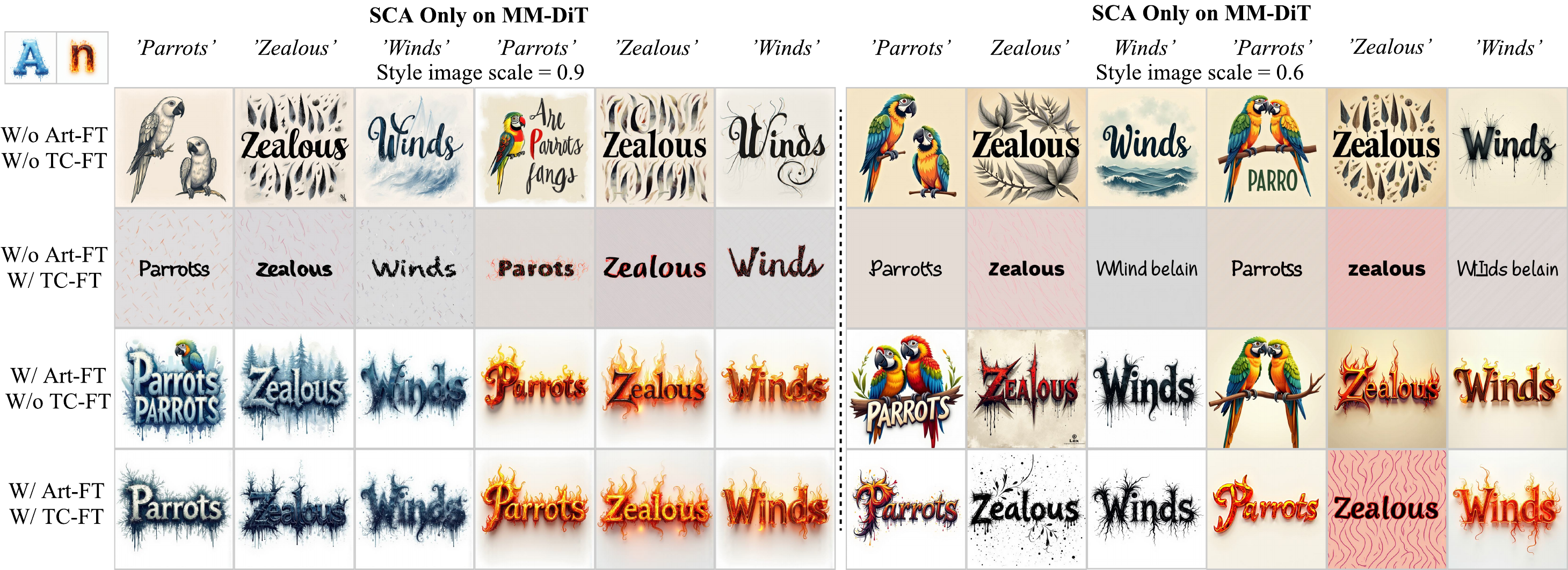}
    \caption{Ablation study of SCA only on MM-DiT with Art-FT and TC-FT.}
    \label{fig:supp-mmdit-only}
\end{figure*}

\begin{figure}[htbp]
  \centering
   \includegraphics[width=\linewidth]{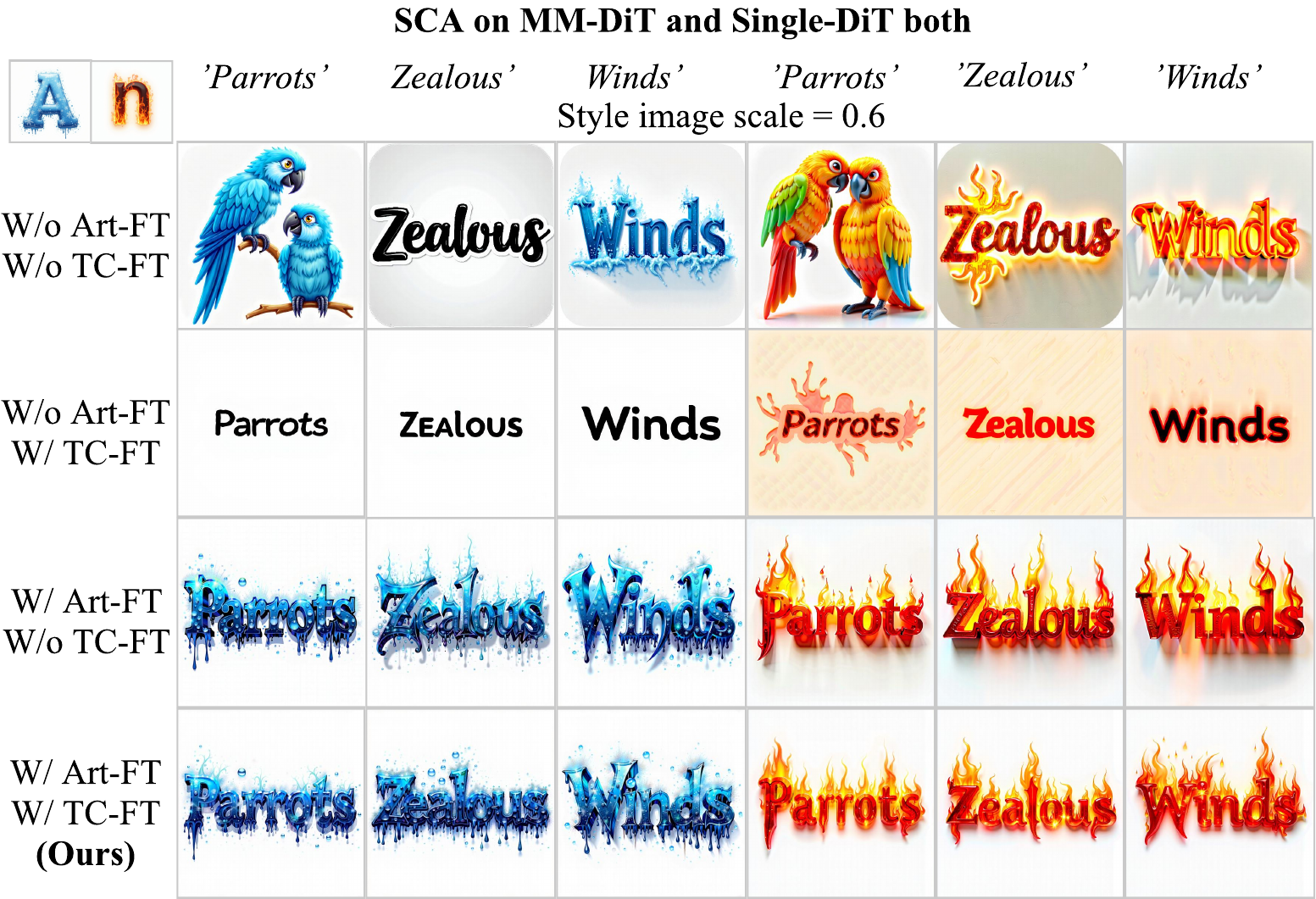}
   \caption{Ablation study of SCA on MM-DiT and Single-DiT both, with Art-FT and TC-FT when image scale = 0.6.}
   \label{fig:ablation_sca_TCFT_0.6}
\end{figure}

\begin{figure}[htbp]
\centering
    \includegraphics[width=\linewidth]{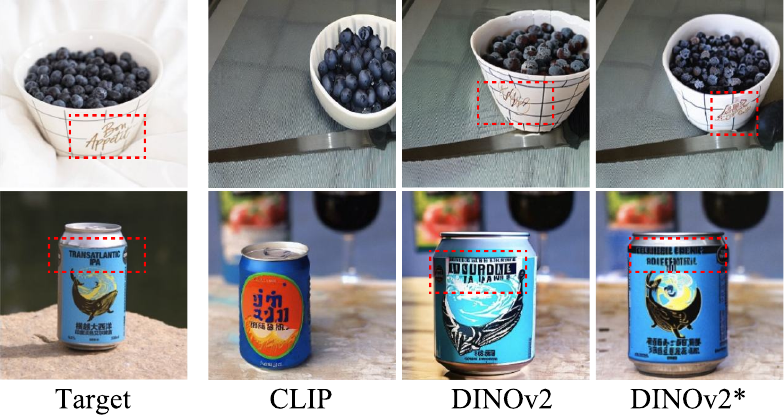}
    \caption{Results of different backbones for the ID extractor in AnyDoor~\cite{chen2024anydoor}. “DINOv2*” refers to removing the background of the target object with a frozen segmentation model before feeding it into the DINOv2 model. This figure is adapted from~\cite{chen2024anydoor}.}
    \label{fig:supp-clip}
\end{figure}

\begin{figure*}[htbp]
\centering
    \includegraphics[width=\linewidth]{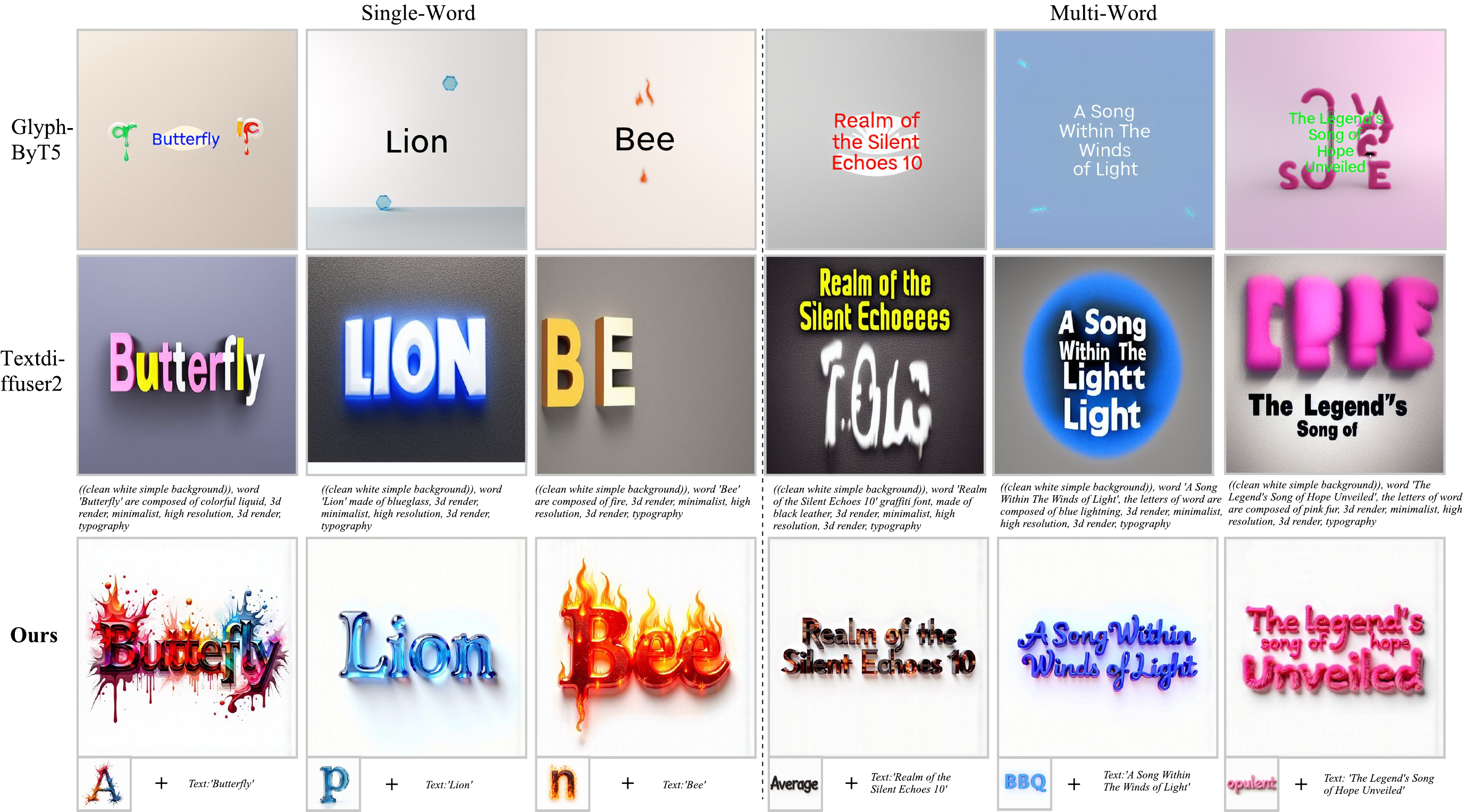}
    \caption{Results of Glyph-ByT5 \cite{liu2024glyph} and Textdiffuser-2 \cite{chen2024textdiffuser} on ATR-bench.}
    \label{fig:supp-gt-atr}
\end{figure*}

\begin{figure*}[htbp]
\centering
    \includegraphics[width=\linewidth]{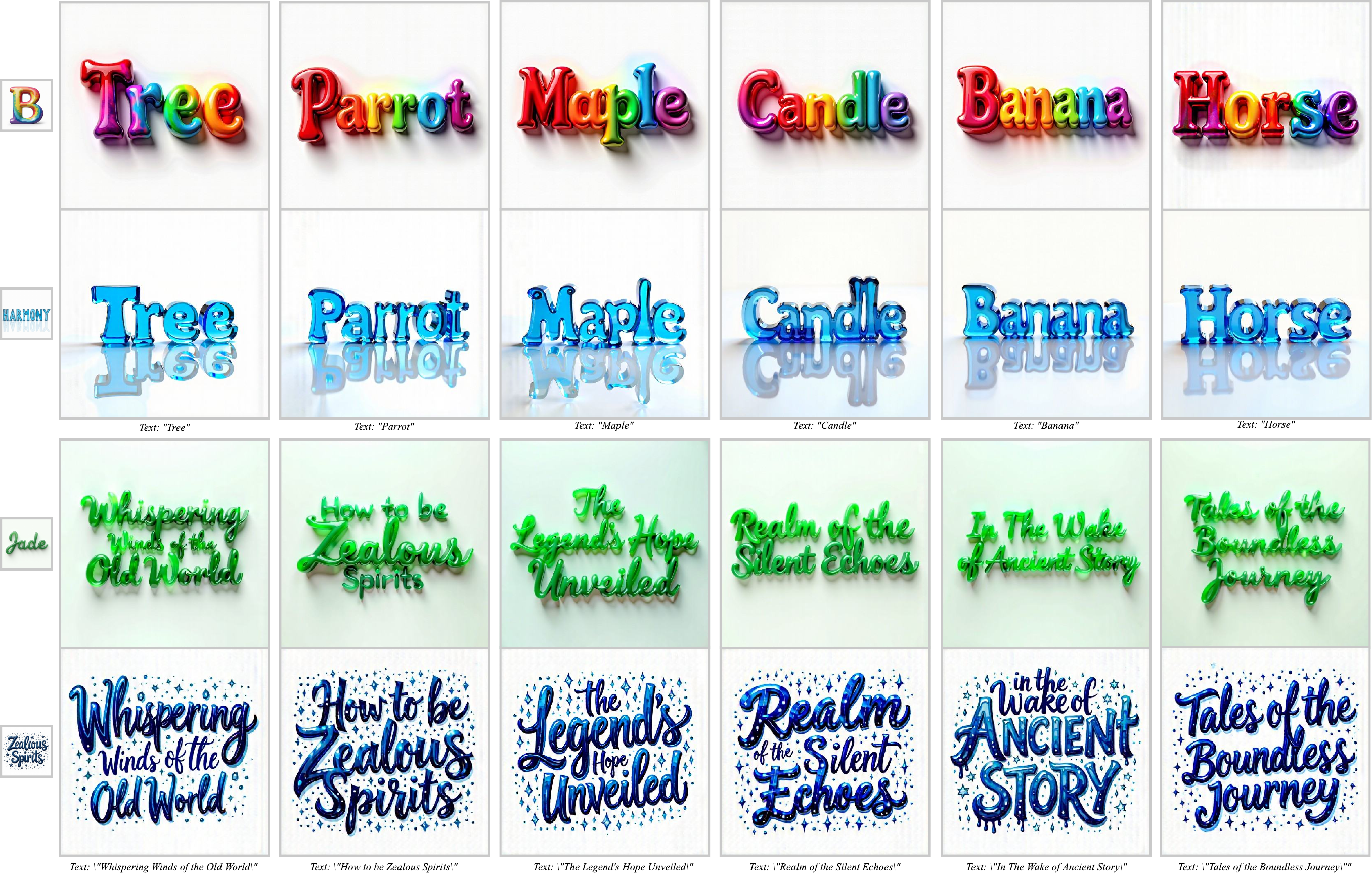}
    \caption{More qualitative results of ours on artistic text rendering.}
    \label{fig:supp-more-atr}
\end{figure*}

\begin{figure*}[htbp]
\centering
    \includegraphics[width=0.99\linewidth]{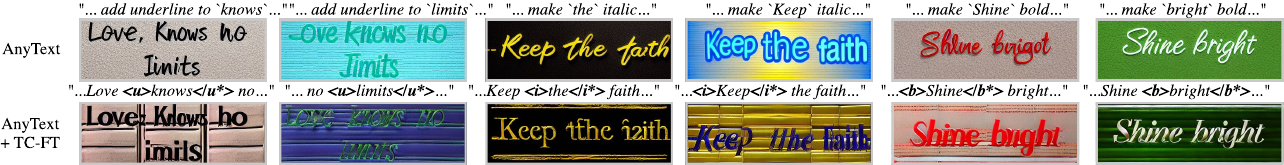}
    \caption{Qualitative results of AnyText~\cite{anytext24} and with TC-Finetuned on BTR.}
    \label{fig:anytext-qual}
\end{figure*}

\begin{figure}[htbp]
  \centering
   \includegraphics[width=\linewidth]{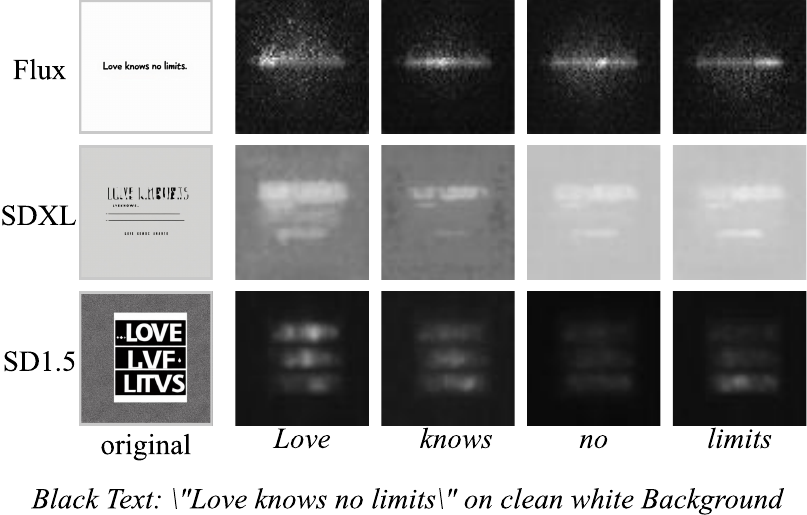}
   \vspace{-2mm}
   \caption{The visualization of attention map on each word in different base models.}
   \label{fig:attn_map}
   \vspace{-2mm}
\end{figure}

\begin{figure}[htbp]
  \centering
   \includegraphics[width=\linewidth]{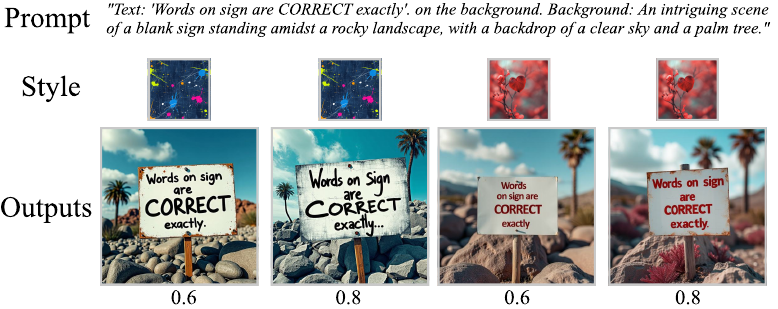}
   \caption{The results of stylized scene text image with different images and image scales.}
   \label{fig:app-1}
\end{figure}

\begin{figure*}[htbp]
\centering
    \includegraphics[width=0.99\linewidth]{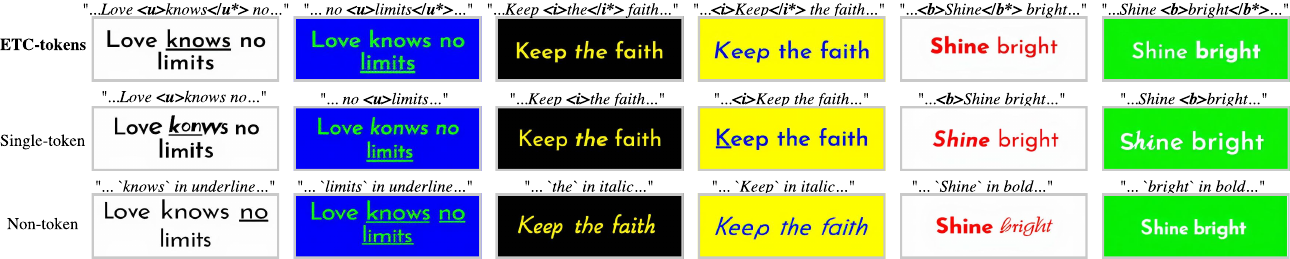}
    \caption{Visual results of ablation on ETC-tokens.}
    \label{fig:ablation-qual}
\end{figure*}

\begin{figure*}[htbp]
\centering
    \includegraphics[width=0.99\linewidth]{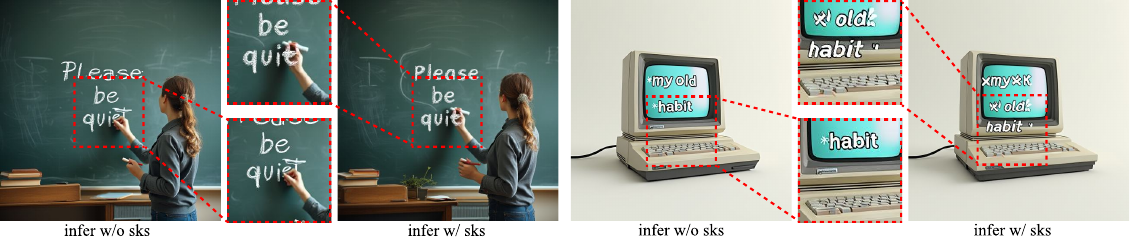}
    \caption{Infer without `sks' to mitigate scene-text detachment.}
    \label{fig:sks}
\end{figure*}

\begin{figure*}[htbp]
  \centering
   \includegraphics[width=0.85\linewidth]{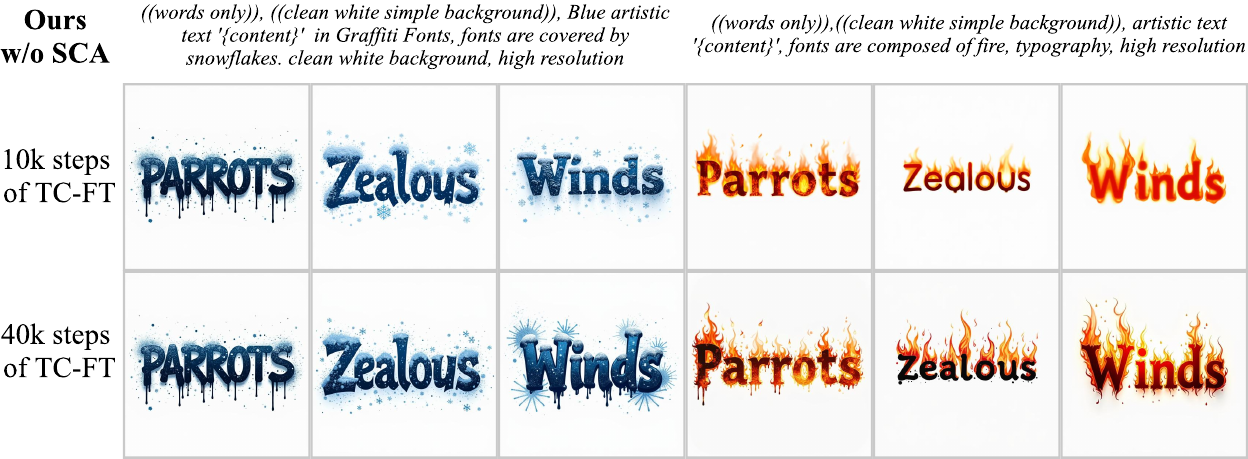}
   \caption{Ablation study of style control adapter (SCA), results from style captions only after 10k and 40k steps of TC-finetuning.}
   \label{fig:ablation_sca}
\end{figure*}

\begin{figure*}[htbp]
\centering
    \includegraphics[width=\linewidth]{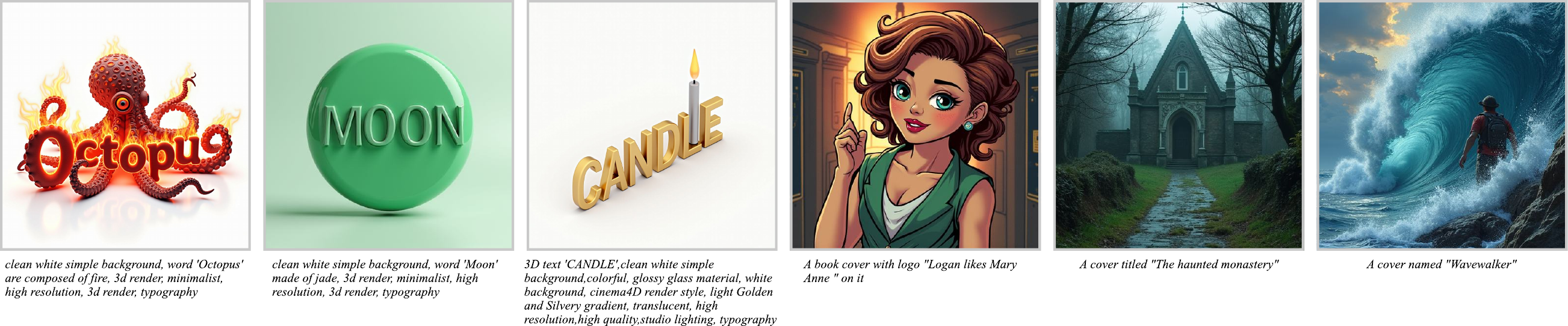}
    \caption{Examples of Semantic Confusion in Flux.1-dev ~\cite{blackforestlabs2024}. The prompts for the right three images are from MARIO-bench~\cite{chen2023textdiffuser}.}
    \label{fig:supp-semantic}
\end{figure*}

\begin{figure*}[htbp]
\centering
    \includegraphics[width=\linewidth]{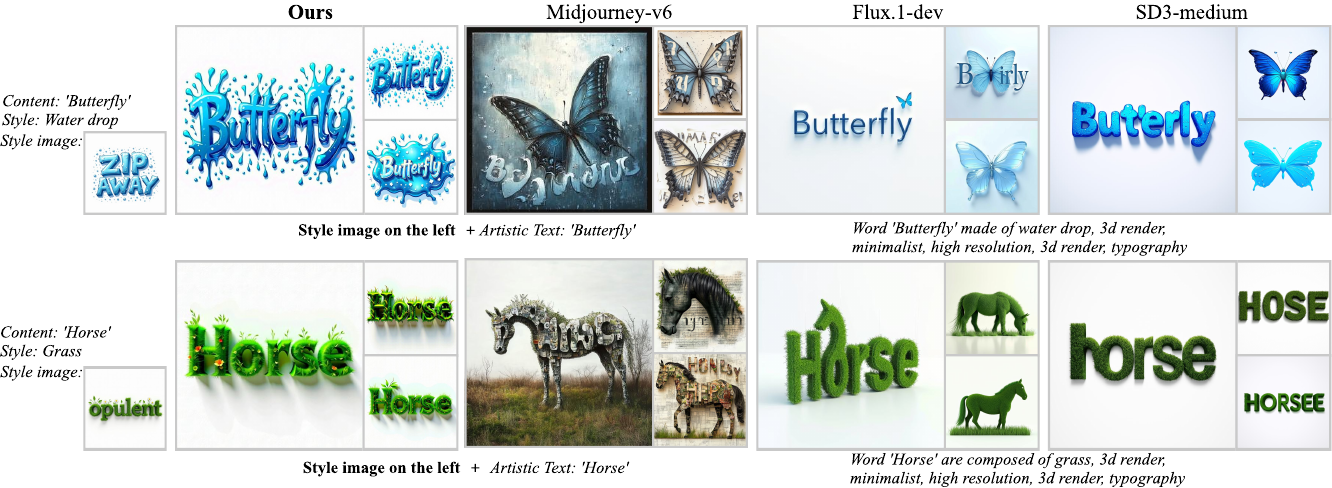}
    \vspace{-2mm}
    \caption{Semantic confusion can also be observed in SD3, Flux and Midjourney.}
    \vspace{-2mm}
    \label{fig:supp-semantic-2}
\end{figure*}

\begin{figure*}[htbp]
\centering
    \includegraphics[width=1\linewidth]{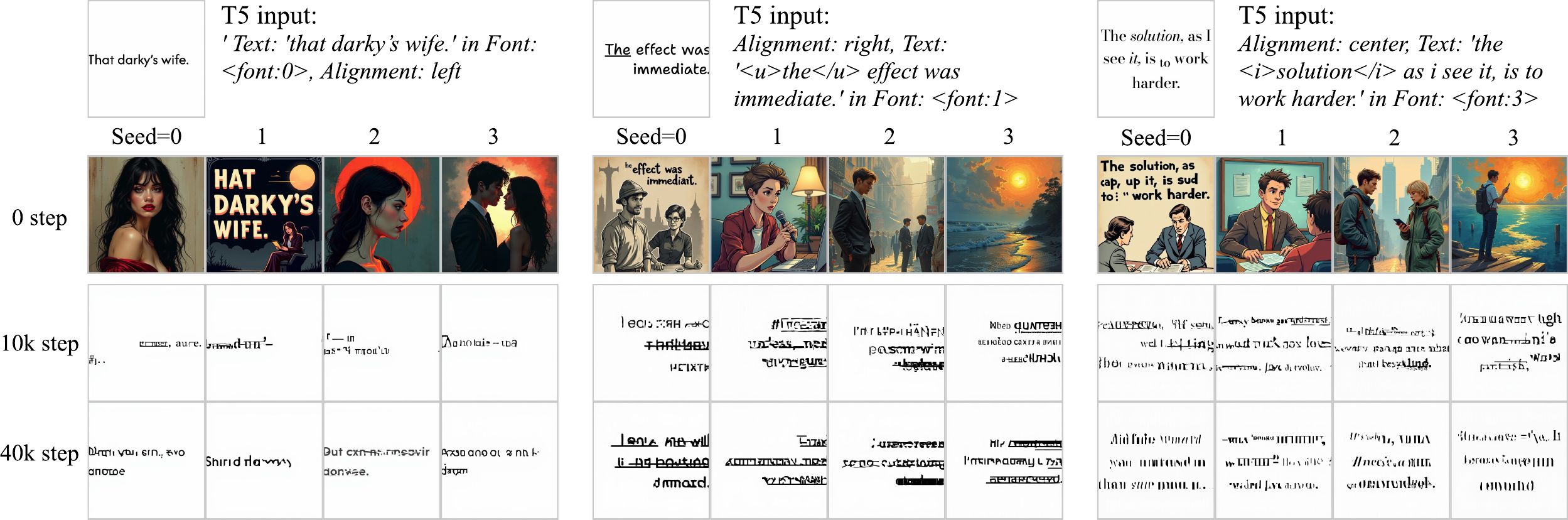}
    \vspace{-2mm}
    \caption{Results of fine-tuning T5 text encoder with new tokens, while input for CLIP is fixed prompt: `words only, clean background'.}
    \vspace{-2mm}
    \label{fig:supp-FT-T5}
\end{figure*}

\begin{figure*}[htbp]
\centering
   \includegraphics[width=\linewidth]{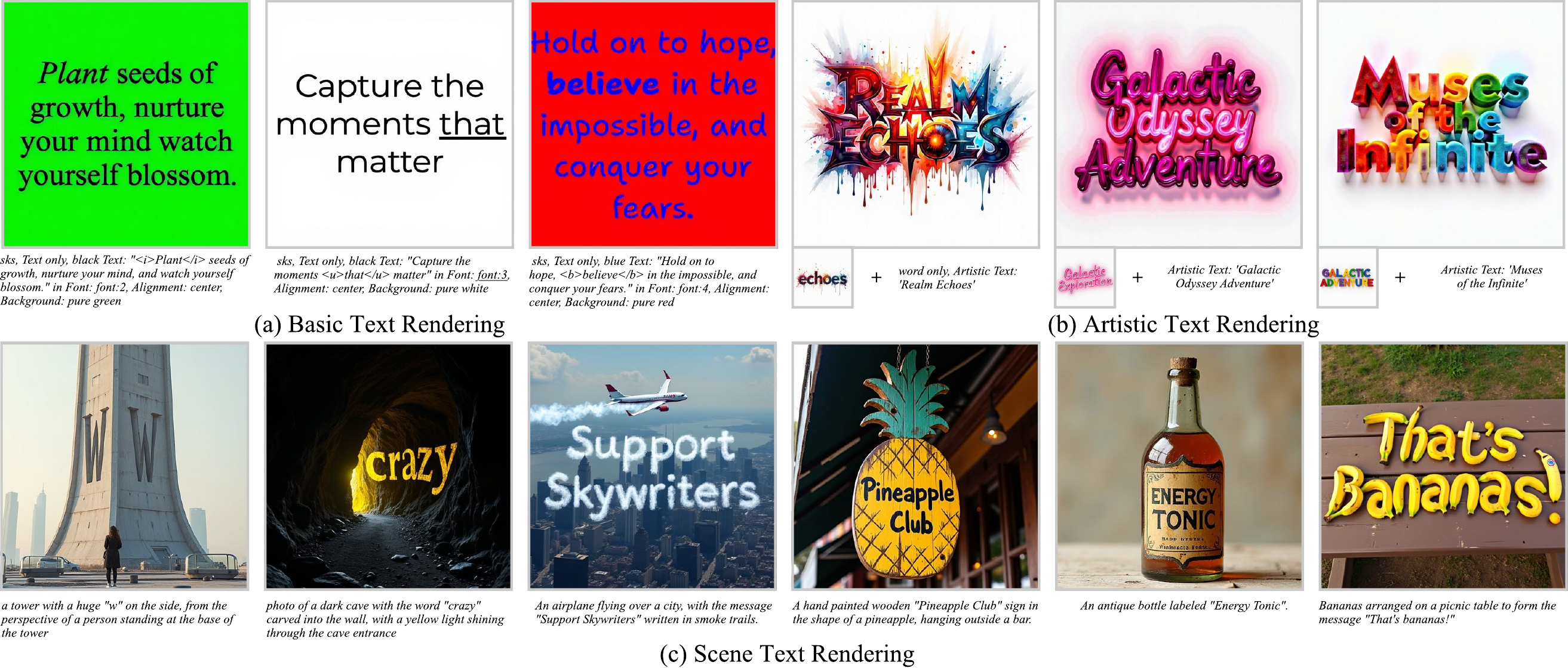}
   \vspace{-2mm}
   \captionof{figure}{Results of our method: (a), (b), and (c) in basic text rendering, artistic text rendering, and scene text rendering, respectively.}
   \vspace{-2mm}
   \label{fig:supp-1}
\end{figure*}

\begin{figure*}[htbp]
\centering
    \includegraphics[width=\linewidth]{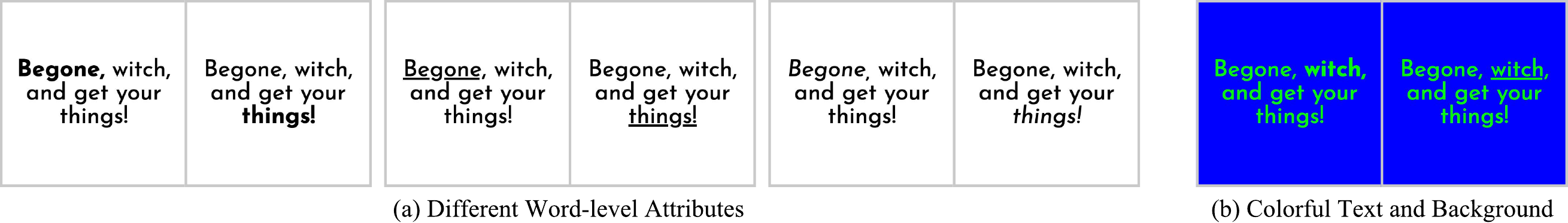}
    \vspace{-2mm}
    \caption{Examples of TC-Dataset. (a) different word-level attributes, (b) examples featuring text and background color variations.}
    \vspace{-2mm}
    \label{fig:supp-2}
\end{figure*}

\begin{figure*}[htbp]
\centering
    \includegraphics[width=\linewidth]{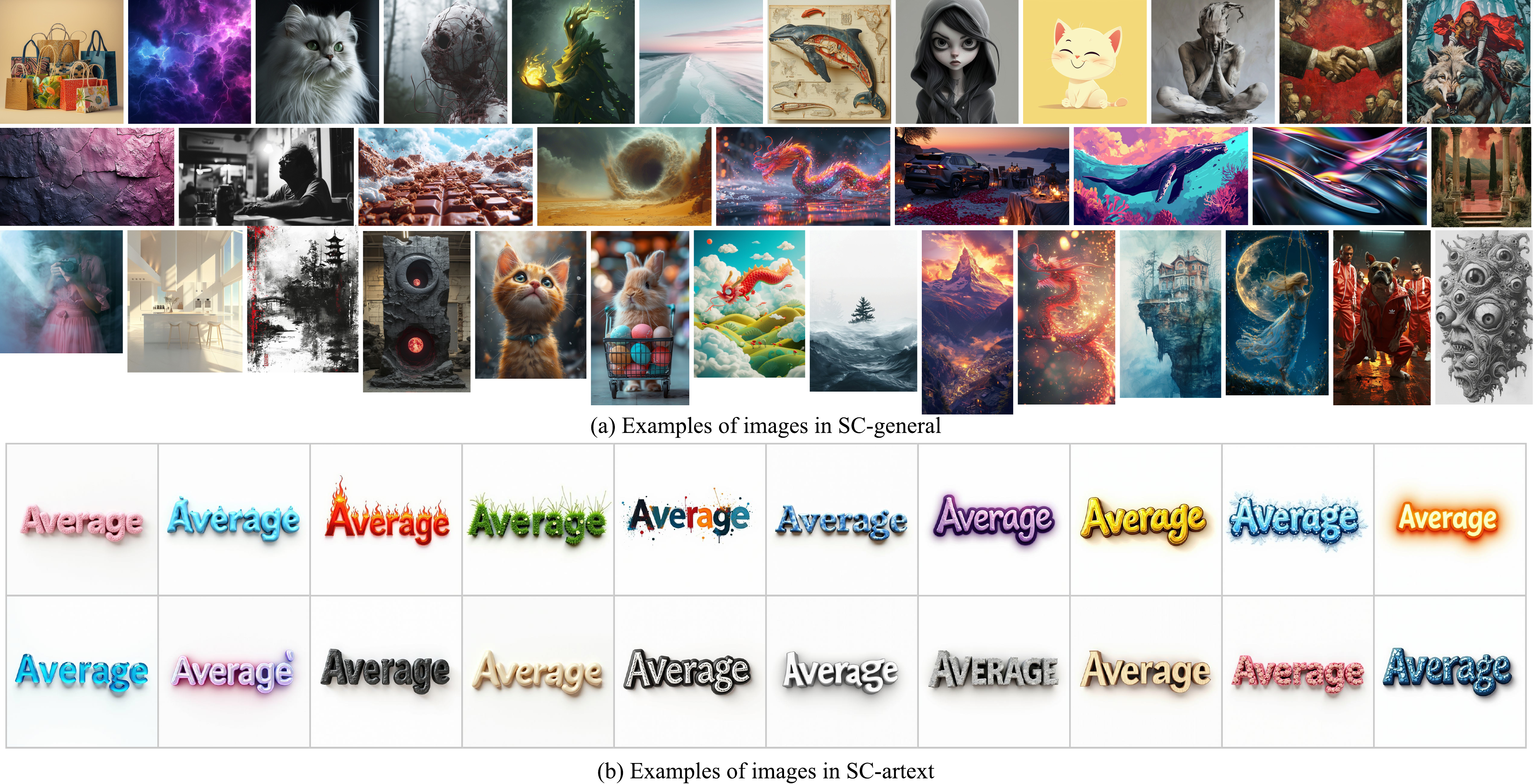}
    \vspace{-2mm}
    \caption{Examples of images in SC-dataset, (a) is SC-general, and (b) is SC-artext.}
    \vspace{-2mm}
    \label{fig:supp-3}
\end{figure*}

\begin{figure*}[htbp]
\centering
    \includegraphics[width=0.99\linewidth]{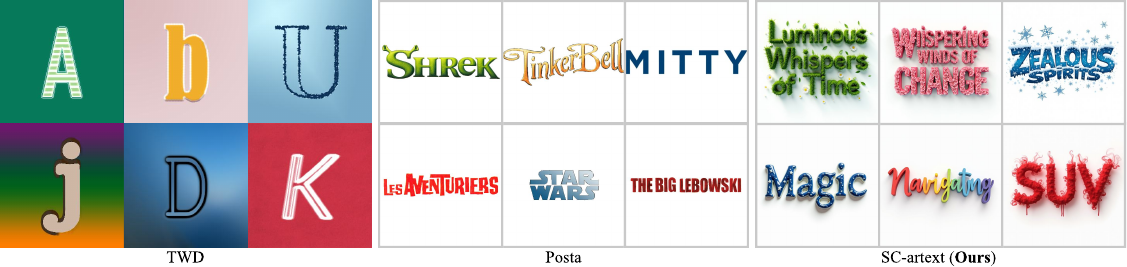}
    \vspace{-3mm}
    \caption{Visual comparison of existing artistic text datasets.}
    \vspace{-5mm}
    \label{fig:compare-data}
\end{figure*}

\noindent{}This supplementary material serves as a complement to the main paper, including additional results presented in Section~\ref{sec1}; more ablation studies of TC-FT, ETC-tokens, and SCA detailed in Section~\ref{sec2}; demonstration of BTR, ATR and STR in Section~\ref{sec3}; discussion of semantic confusion is detailed in Section~\ref{sec4}; details of the datasets used in Section~\ref{sec5}; details of word accuracy (Word-Acc) in Section~\ref{sec6}; and further details regarding user study in Section~\ref{sec7}.

\section{More Results}
\label{sec1}

\subsection{Typographic Controls in STR}
We found that the typography controls acquired from Basic Text Rendering (BTR) can be partially transferred to other text rendering tasks. The model's capacity to learn typography attributes from simple text images shows considerable promise for generalization and adaptability in various domains. Consequently, this enables the application of typographic controls, as depicted in Figure \ref{fig:supp-con-STR}, and font selection, as displayed in Figure \ref{fig:supp-font-STR}, in Scene Text Rendering (STR).

\subsection{Differences with Flux-IPA}

1) Our style control adapters (SCA) employ a two-stage training approach. Fine-tuning with SC-artext significantly boosts artistry without compromising the accuracy of text, making it more suitable for the ATR task.  

\noindent{}2) In contrast to Flux-IPA(XLabs) \footnote{\href{https://huggingface.co/XLabs-AI/flux-ip-adapter}{Flux-IPA(XLabs)}}, which is only applied on MM-DiT, our SCA is implemented on both MM-DiT and Single-DiT to enhance style control, as depicted in Figure~\ref{fig:supp-mmdit-only} with Figure~\ref{fig:ablation_sca_TCFT_0.6}. Even with a style image scale of 0.6, the style achieved by applying SCA on both MM-DiT and Single-DiT is markedly superior to that of applying SCA only on MM-DiT with a style image scale of 0.9. The comparison between Table~\ref{tab:ablation-SCA-both} and Table~\ref{tab:ablation-SCA-only} further validates this, as applying SCA on both MM-DiT and Single-DiT yields a higher CLIP-I score under different settings. 

\noindent{}3) Unlike Flux-IPA(InstantX) \footnote{\href{https://huggingface.co/InstantX/FLUX.1-dev-IP-Adapter}{Flux-IPA(InstantX)}} which uses SigLIP~\cite{zhai2023sigmoid}, our method select CLIP~\cite{radford2021learning} as the image encoder. This choice is grounded in the distinct characteristics of these two models. SigLIP~\cite{zhai2023sigmoid, tschannen2025siglip} is renowned for its robust OCR capabilities. Conversely, as discussed in \cite{liu2024glyph, chen2024anydoor}, CLIP's visual embeddings are insensitive to text. This insensitivity to text in CLIP's visual embeddings is pivotal for our application, as it mitigates content leakage from style images (artistic text images). The visual outcomes presented in Figure~\ref{fig:supp-clip} provide empirical evidence in support of our selection.

\noindent{}4) Distinct from previous methods, we insert adapters in an interval-skip manner (on layer 0,2,4...) to reduce costs. In terms of parameter usage, the parameters of adapters in Flux-IPA(InstantX) are approximately 2.85 times ours, as demonstrated in Table~\ref{tab:compare-para}.

\subsection{More Qualitative Results of ATR}
This section serves as a supplement to Section {\color{iccvblue} 4.2} of the main paper, offering a qualitative comparison of our method with Glyph-ByT5 \cite{liu2024glyph} and Textdiffuser-2 \cite{chen2024textdiffuser} on the ATR-bench dataset, as depicted in Figure \ref{fig:supp-gt-atr}. Additionally, we present our extended qualitative results on the ATR-bench dataset, including single-word and multi-word examples, in Figure~\ref{fig:supp-more-atr}. Notably, in the second row of results in Figure~\ref{fig:supp-more-atr}, the accurate mirror reflection of letters in every result further substantiates the effectiveness of our SCA. This example showcases that our SCA can inject style while meticulously maintaining text accuracy, providing additional empirical evidence for the capabilities of our proposed approach in the ATR task.

\subsection{Train Baseline}
In addition to the aforementioned comparisons, we fine-tune another baseline, AnyText~\cite{anytext24} on the TC-dataset using a method similar to TC-finetuning. The quantitative results are presented in Table~\ref{tab:compare-quant-btr}, while the qualitative results are shown in Figure~\ref{fig:anytext-qual}. These results clearly reveal that AnyText fails to acquire word-level controllability. The performance of Glyph-ByT5 and Textdiffuser-2 exhibits similar limitations. This may be attributed to the inherent restricted capabilities of the base models for text rendering. Figure~\ref{fig:attn_map} shows the attention maps of different base models for different words in basic text rendering.

\subsection{Stylization of STR} 
With our SCA, the influence of style input on the text within the image is minimal, as clearly observable in Figure~\ref{fig:app-1}. When distinct style images are incorporated, a pronounced transformation in the text style ensues. Notwithstanding these changes in style, the integrity of the text content is maintained, remaining accurate and distinguishable.

\section{More Ablation}
\label{sec2}

\subsection{Ablation on SCA}
\textbf{SCA Only on MM-DiT.} Upon comparing Figure~\ref{fig:supp-mmdit-only} and Figure~\ref{fig:ablation_sca_TCFT_0.6}, it is observed that when SCA is implemented on both MM-DiT and Single-DiT, the degree of stylization achieved is substantially greater than when SCA is applied solely to MM-DiT. This holds true even when the scale of the style image is lower (images Figure~\ref{fig:ablation_sca_TCFT_0.6}) in the former case (left images in Figure~\ref{fig:supp-mmdit-only}). A comparison between Table~\ref{tab:ablation-SCA-both} and Table~\ref{tab:ablation-SCA-only} provides additional validation of this assertion when evaluated in the context of CLIP-I metrics.

\noindent{}\textbf{SCA with Art-FT and TC-FT.} The CLIP-I and OCR-Acc presented in Table~\ref{tab:ablation-SCA-both} are the average figures obtained on ATR task when the scale of the style image is set at 0.9 and 0.6, respectively. Table~\ref{tab:ablation-SCA-both} is identical to Table 6 in the main paper. These values are placed here to enable a more direct comparison with SCA only on MM-DiT (Table~\ref{tab:ablation-SCA-only}). It becomes evident that, irrespective of whether SCA, the impacts of Art-FT and TC-FT on the ATR task remain consistent: Art-FT enhances stylization, while TC-FT improves content accuracy. Additionally, as shown in Table~\ref{tab:compare-delta}, after Art-FT, the degree of style degradation caused by TC-FT is reduced. This highlights the distinct but complementary roles of Art-FT and TC-FT in optimizing both the stylistic and content-related aspects of the results.

\noindent{}\textbf{Without SCA.} As is evident from Figure~\ref{fig:ablation_sca}, in the absence of SCA, even when a detailed style caption is employed to characterize the style, diverse text contents result in inconsistent styles under the same random seed. Moreover, through a comparison of the images in the two rows, it becomes apparent that TC-FT exerts a certain degrading impact on the artistic style imparted by the style caption. 

\subsection{Ablation on TC-FT}
Regarding the ablation study of typography control fine-tuning (TC-FT), we configured four distinct training scenarios: (1) only new tokens, (2) T5 text encoder with new tokens, (3) joint text attention (Txt-Attn) with new tokens, and (4) joint text-image attention (Txt+Img-Attn) with new tokens. As previously established in \cite{esser2024scaling, liu2024glyph}, text rendering performance is primarily governed by the text encoder architecture. To explore this, we attempted to fine-tune the T5 on the BTR dataset to enhance controllability in text rendering. However, this approach led to a substantial decline in text accuracy, with visual artifacts evident in the generated outputs. The visual results are documented in Figure~\ref{fig:supp-FT-T5}.

\subsection{Ablation on ETC-Tokens}
This section supplements Section {\color{iccvblue} 4.3} of the main paper, focusing on demonstrating the effectiveness of the proposed Enclosing Typography Control (ETC)-tokens for targeted word-level typographic attributes. For instance, to bold the word "robot" in the phrase ``i am not a robot", we explore three settings: 
1) Non-Token: Using an instruction prompt instead of adding modifier tokens, such as ``the `robot' is in bold". 2) Single-Token: Following~\cite{kumari2023multi,butt2025colorpeel}, we trained our model to use a single token, placing the modifier token before ``robot". 3) Our ETC-Token. The visual results of ablation on ETC-tokens are presented in Figure~\ref{fig:ablation-qual}.

\subsection{Ablation on `sks' prefix}
We tried to mitigate language drift (scene-text detachment) by training with the `sks' prefix in prompts of the TC-Dataset, which is omitted during inference. This low-cost approach helps alleviate detachment, as shown in Figure~\ref{fig:sks}.


\begin{table}[ht]
\centering
\scalebox{1}{
\begin{tabularx}{\columnwidth}{c c c c c}
\toprule
Art-FT & TC-FT & CLIP-I ${\uparrow}$ &  OCR-Acc ${\uparrow}$ &  Avg ${\uparrow}$\\
\midrule
\ding{55} & \ding{55} & 60.07 & 28.89 & 44.48 \\
\ding{55} & \ding{51} & 58.09 & \textbf{65.39}  & 61.74 \\
\ding{51} & \ding{55} & \textbf{65.12} & 34.48 & 49.80 \\
\ding{51} & \ding{51} & 64.27 & 60.07 & \textbf{62.17} \\
\bottomrule 
\end{tabularx}}
\caption{Ablation studies of fine-tuning with SC-artext (Art-FT) for SCA (on MM-DiT and Single-DiT both) and TC-finetuning (TC-FT) for backbone. The last row is ours.}
\label{tab:ablation-SCA-both}
\end{table}

\begin{table}[ht]
\centering
\scalebox{1}{
\begin{tabularx}{\columnwidth}{c c c c c}
\toprule
Art-FT & TC-FT & CLIP-I ${\uparrow}$ &  OCR-Acc ${\uparrow}$ &  Avg ${\uparrow}$\\
\midrule
\ding{55} & \ding{55} & 54.19 & 24.32 & 39.26 \\
\ding{55} & \ding{51} & 51.64 & \textbf{60.79}  & 56.22 \\
\ding{51} & \ding{55} & \textbf{58.14} & 17.89 & 38.02 \\
\ding{51} & \ding{51} & 56.40 & 58.27 & \textbf{57.34} \\
\bottomrule 
\end{tabularx}}
\caption{Ablation studies of fine-tuning with SC-artext (Art-FT) for SCA (only on MM-DiT) and TC-finetuning (TC-FT) for backbone.}
\label{tab:ablation-SCA-only}
\end{table}

\begin{table}[t]
    \centering
    \begin{tabular}{m{2cm}m{2cm}m{1.8cm}}
        \toprule
        Modules  & Non-Skip & Skip (Ours) \\
        \midrule
        Adapters & 1434.45 M & 503.38 M \\
        \bottomrule
    \end{tabular}
      \caption{Parameter quantity comparison with Flux-IPA(InstantX).}
      \label{tab:compare-para}
\end{table}

\begin{table}[t]
    \centering
    \begin{tabular}{m{2cm}m{2.5cm}m{2.5cm}}
        \toprule
        $\Delta_{CLIP-I}$ & w/o Art-FT & w/ Art-FT \\
        \midrule
        Both & 1.98 \tiny($60.07\to58.09$) & 0.85 \tiny ($65.12\to64.27$) \\
        Only & 2.55 \tiny($54.19\to51.64$) & 1.74 \tiny($58.14\to56.40$) \\
        \bottomrule
    \end{tabular}
      \caption{Comparison of CLIP-I changes with and without Art-FT in two SCA settings after TC-FT. Both: SCA on MM-DiT and Single-DiT both, Only: SCA only on MM-DiT.}
      \label{tab:compare-delta}
\end{table}

\begin{table}[t]
    \centering
    \begin{tabular}{m{2.4cm}m{1.4cm}m{1.5cm}m{1.4cm}}
        \toprule
        Methods & \footnotesize OCR-Acc ${\uparrow}$ & \footnotesize Word-Acc ${\uparrow}$ & \footnotesize Font-Con ${\uparrow}$ \\
        \midrule
        AnyText  & 43.78  & \ding{55}  & 3.67  \\
        AnyText \footnotesize{+TC-FT} & 39.26 & \ding{55} & 2.64 \\
        \textbf{Ours} & \textbf{82.85} & \textbf{55.00} & \textbf{68.42} \\
        \bottomrule
    \end{tabular}
      \caption{Quantitive results of AnyText and with TC-FT on BTR.}
      \label{tab:compare-quant-btr}
\end{table}

\begin{table}[t]
\centering
\scalebox{1}{
\begin{tabular}{m{3cm}m{2cm}m{2cm}}
\toprule
Methods & OCR-Acc${\uparrow}$ &  Word-Acc${\uparrow}$ \\
\midrule
Non-Token & 71.00 & 25.00  \\
Single-Token & 72.88 & 32.00 \\
ETC-Token(\textbf{Ours}) & \textbf{82.85} & \textbf{55.00} \\
\bottomrule 
\end{tabular}}
\caption{Ablation studies of ETC-Token on basic text rendering.}
\label{tab:ablation-etc}
\end{table}

\section{Demonstration of BTR, ATR and STR}
\label{sec3}

This section provides additional information to complement Section {\color{iccvblue} 1} of the main paper, which outlines the scope of three text rendering tasks:

\begin{itemize}
    \item Basic Text Rendering (BTR) involves rendering simple text on a solid color background without any additional scene elements, as illustrated in Figure \ref{fig:supp-1}(a). 
    \item Artistic Text Rendering (ATR) features a minimalist background that highlights the artistic nature of the text itself, as seen in Figure \ref{fig:supp-1}(b). 
    \item Scene Text Rendering (STR) involves integrating text and scene elements in a way that shares contextual meaning and blends harmoniously, as depicted in Figure \ref{fig:supp-1}(c). 
\end{itemize}

\section{Semantic Confusion}
\label{sec4}
The term ``semantic confusion" in the main paper refers to instances where text rendering incorrectly generates visual objects based on the semantic meaning of the text, rather than just producing the text itself. For example, as shown in Figure \ref{fig:supp-semantic}, our intention was to render only the artistic text ``Octopus", ``MOON", and ``CANDLE" in the left three images. However, the images inadvertently include the corresponding objects for these words. Similarly, in the right three images, which are supposed to display text on the scene, the text is absent, and only the specific objects associated with the semantic meaning of text are present.

Additionally, we conducted additional comparisons with Midjourney~\cite{mj_website}, Flux, and SD3 in Figure~\ref{fig:supp-semantic-2}. Whereas original SD3 and Flux lack the capability to process image inputs, both our proposed approach and Midjourney demonstrate the ability to handle combined image-text prompts. The results presented in the figure highlight a critical observation: during artistic text rendering tasks, semantic ambiguity significantly impairs the model's capacity to accurately render the specified word's content. Instead, the model tends to generate visual representations corresponding to the word's semantic reference rather than words itself its. This phenomenon underscores the challenges inherent in balancing stylization and content accuracy within artistic text rendering.

\begin{table}
    \scriptsize  
    \centering
    \begin{tabularx}{\columnwidth}{c c c c c c}
        \toprule
        Metrics \textbackslash Dataset & TWD~\cite{typography2019} & Posta~\cite{chen2025posta} & SC-artext & SC-general \\
        \midrule
        Aesthetic ${\uparrow}$ & 42.64 & 47.23 & \underline{68.57} & \textbf{84.78} \\
        Quality ${\uparrow}$ & 55.83 & 64.89 & \underline{88.73} & \textbf{91.20} \\
        \bottomrule
    \end{tabularx}
       \vspace{-3mm}
      \caption{Aesthetic and quality scores comparison.}
      \label{tab:compare-aesthetic}
      \vspace{-3mm}
\end{table}

\section{Details of Datasets}
\label{sec5}
This section complements Section {\color{iccvblue} 3, 4} of the main paper, detailing the datasets we utilized in our work.

\noindent \textbf{Typography Control Dataset (TC-Dataset).} 
To address the lack of high-quality datasets that integrate text with word-level typographic attributes, we developed the TC-Dataset using typography control rendering (TC-Render). This process harnesses HTML rendering to generate images that display typographic features such as various fonts and word-level attributes, including bold, italic and underline.
We initiated our process by extracting 625 text excerpts from novels. For each excerpt, we designed an HTML structure comprising sixteen images: one without typographic attributes and, in five different positions, applied three distinct typographic attributes (shown in Figure \ref{fig:supp-2} (a)). Furthermore, we applied data augmentation techniques by randomly altering the text color and background (shown in Figure \ref{fig:supp-2} (b)). Each HTML structure was rendered with one of five different fonts, resulting in approximately 50k text-image pairs with solid color backgrounds.

\noindent \textbf{Style Control Dataset (SC-Dataset).} 

\noindent \textit{SC-general.} To train our style control adapters, we assembled the SC-general dataset, which includes approximately 580k general image-text pairs with high aesthetic scores. These pairs were sourced from open-source datasets \cite{laionaesthetics, sun2024journeydb}. Figure \ref{fig:supp-3} (a) presents sample images, and Table \ref{tab:prompt-list} displays the corresponding paired texts.

\noindent \textit{SC-artext.} For fine-tuning the style control adapters, we created the SC-artext dataset. We combined a list of 100 style descriptions with a list of 99 words, categorized into three character length groups: 1-15, 16-30, and 30-50. This combination produced a variety of prompts for artistic text images, which served as input for Flux.1-dev \cite{blackforestlabs2024}, yielding around 20k high-quality images. To ensure the images accurately reflected the original text content, we utilized shareGPT4v \cite{chen2023sharegpt4v} to regenerate captions. Figure \ref{fig:supp-3} (b) shows sample images, and Table \ref{tab:prompt-list} presents the paired texts. Besides, we provide quantitative and qualitative comparison results with artistic text in TWD~\cite{typography2019} and Posta~\cite{chen2025posta} in Table~\ref {tab:compare-aesthetic} and Figure~\ref{fig:compare-data}, respectively. We randomly sample 100 images from each and use specialized LMM (Q-Align~\cite{wu2024q}) for quality and aesthetic evaluation.

\section{Details about Word-Acc}
\label{sec6}
\begin{figure*}[htbp]
\centering
    \includegraphics[width=\linewidth]{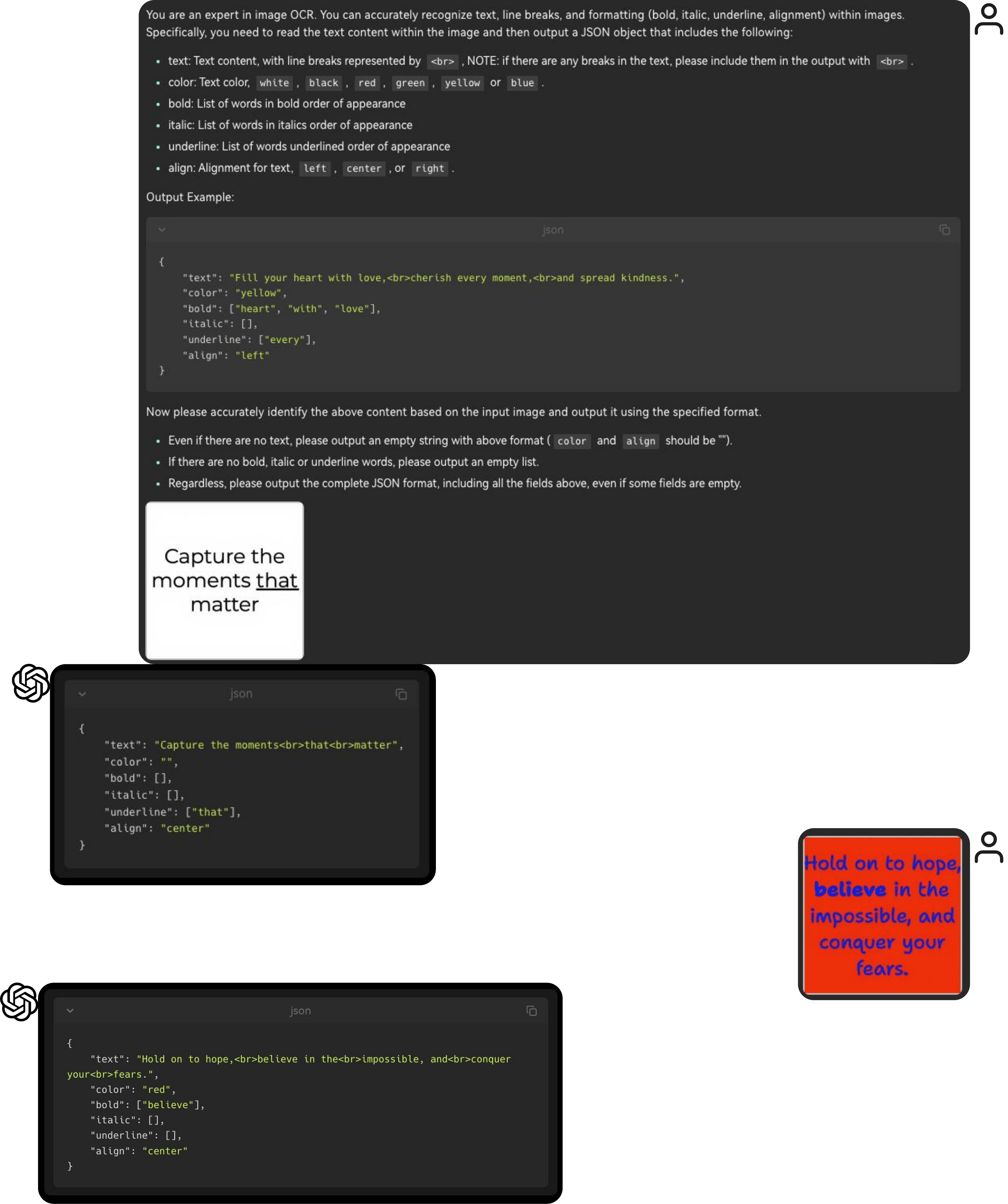}
    \vspace{-5mm}
    \caption{Example of using GPT-4o to evaluate word-level attribute accuracy (Word-Acc).}
    \vspace{-4mm}
    \label{fig:supp-gpt}
\end{figure*}

Current open-source OCR tools lack the capability to recognize word-level attributes such as bold, italic, and underline. To address this limitation, we employ GPT-4o \cite{openai_gpt4o} to evaluate the accuracy of word-level attributes (Word-Acc). We have designed a structured prompt, supplemented with example cases, to improve GPT-4o's precision in predicting these attributes. Figure \ref{fig:supp-gpt} illustrates a dialogue record that showcases GPT-4o's strong context comprehension and logical reasoning abilities.

\section{Details of User Study}
\label{sec7}
\begin{figure*}[htbp]
\centering
    \includegraphics[width=0.8\linewidth]{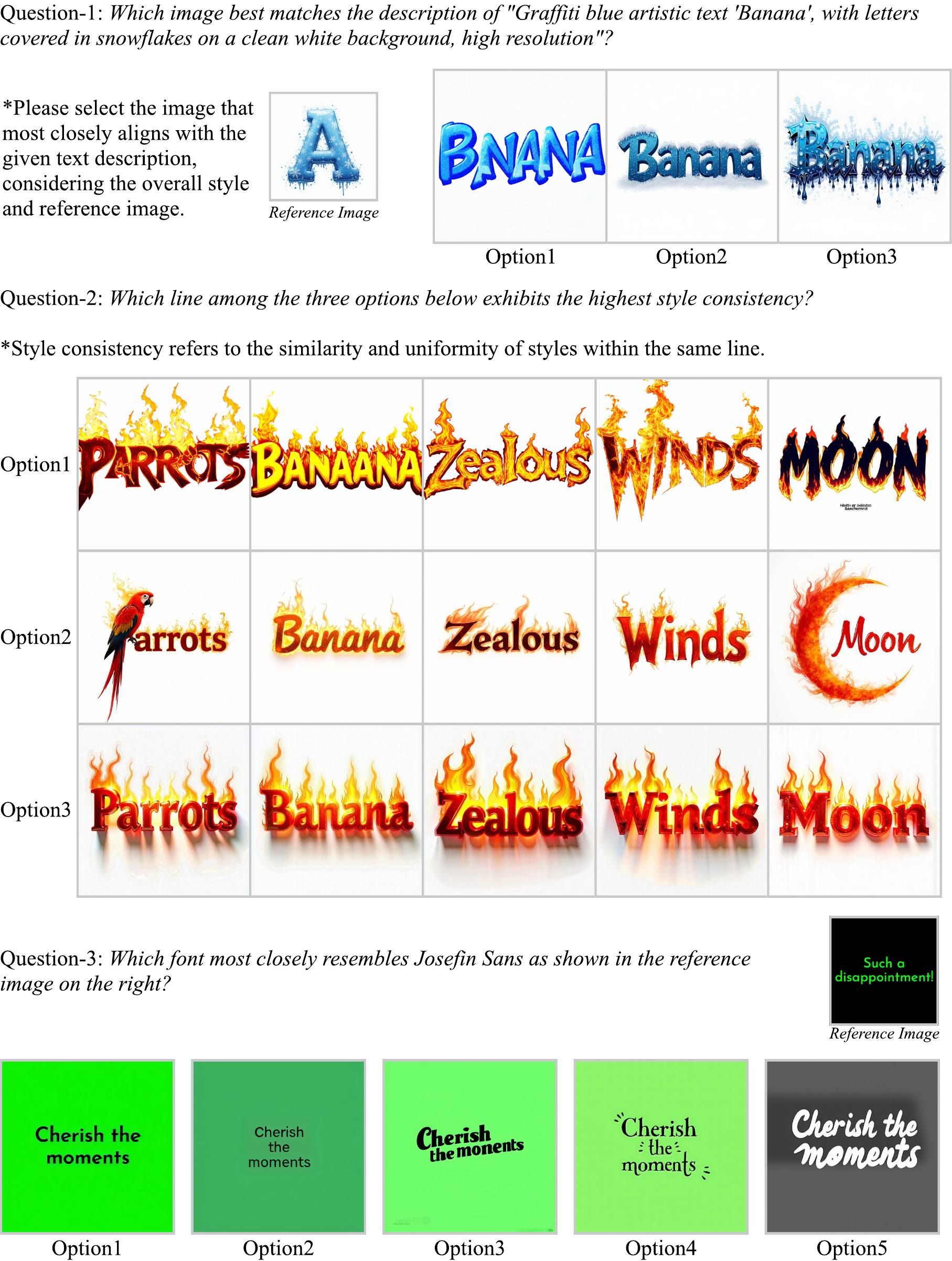}
    \vspace{-2mm}
    \caption{Examples of questionnaire to evaluate the Style-Con and Font-Con.}
    \vspace{-4mm}
    \label{fig:supp-user1}
\end{figure*}

\begin{figure*}[htbp]
\centering
    \includegraphics[width=0.8\linewidth]{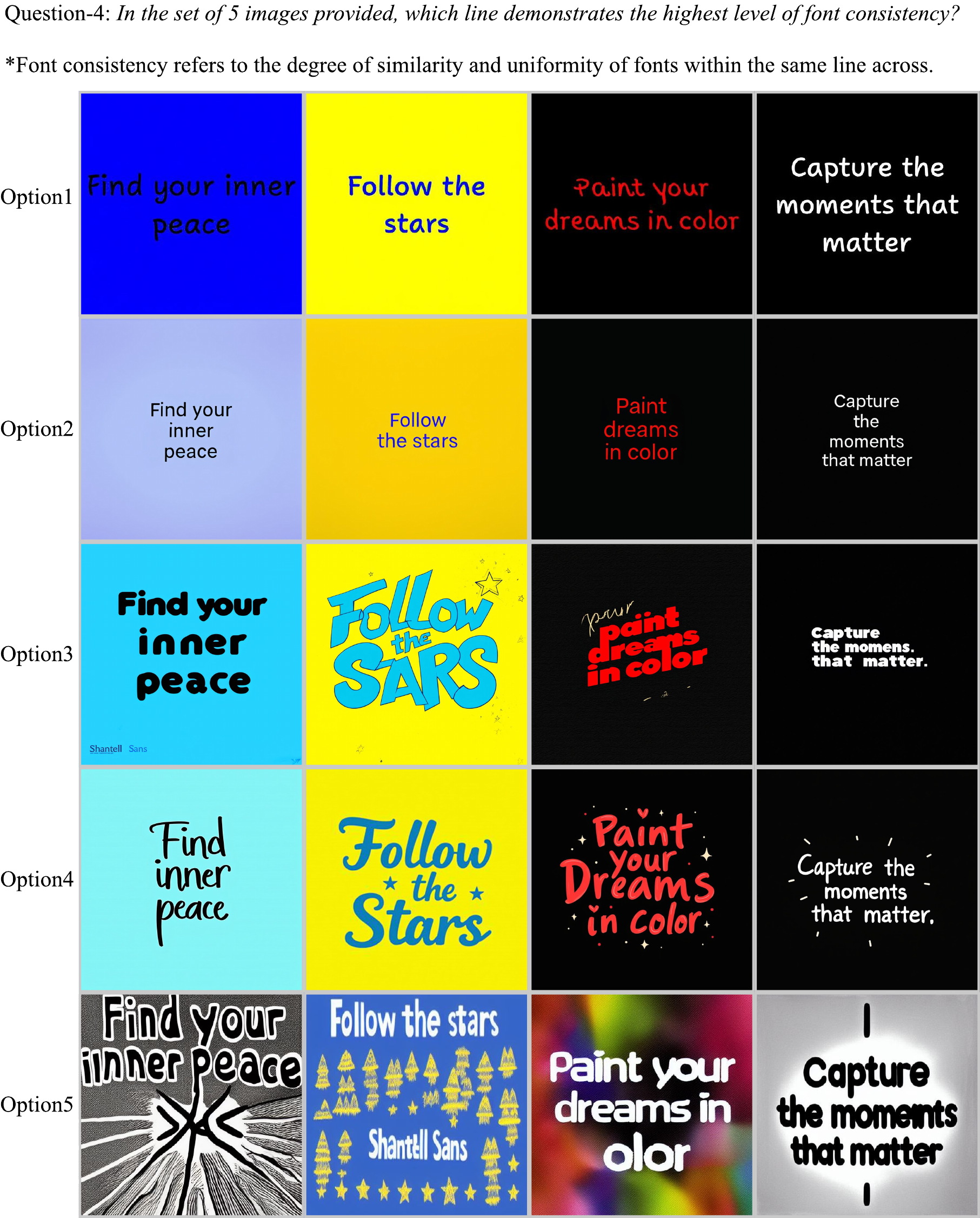}
    \vspace{-2mm}
    \caption{Examples of questionnaire to evaluate the Font-Con.}
    \vspace{-4mm}
    \label{fig:supp-user2}
\end{figure*}

This section complements Section {\color{iccvblue} 4.1} of the main paper, providing additional details on the user studies. We involved 22 participants in these studies to evaluate our results perceptually, comparing them to baseline methods. The evaluation focused on two main aspects: font consistency (Font-Con) and style consistency (Style-Con). For Font-Con, we had two subtypes. One evaluated the consistency between the output image and the ground truth, and the other judged font consistency across different outputs with the same input. Style-Con was evaluated in a similar way, also with two subtypes. Style-Con was evaluated in two ways: one subtype measured the consistency between the output image and the ground truth, while the other assessed the consistency of fonts across different outputs when the same font input was used. This can be seen in Questions 1 and 2 in Figure \ref{fig:supp-user1}. Font-Con was evaluated in a similar manner, with two subtypes addressing the same two aspects. These are represented by Question 3 of Figure \ref{fig:supp-user1} and Question 4 of Figure \ref{fig:supp-user2}. Each subtype had a different number of questions: 4, 2, 3, and 2, respectively. The score for each method was determined by dividing the number of votes it received by the total number of votes cast.


\begin{table*}[t] 
\centering
\begin{tabularx}{\textwidth}{p{2cm}|X}
\toprule
Image & Text \\
\midrule
SC-general, Row 1, Col 1 & A photorealistic image of multiple shopping bags in a boho style, fresh and inviting. The bags are in various sizes and patterns, including floral designs, abstract prints, and earthy tones. They have rope handles and are arranged against a soft, neutral background. The overall vibe is natural, stylish, and vibrant. \\
\hline
SC-general, Row 1, Col 2 & Dark blue purple red abstract background for design. Painted rough paper. Bright colors include magenta and fuchsia. Smudge, stain, and blot effects are photo-realistic with ultra sharp focus and ultra detailed focus. The image has high coherence and minimalistic style with intricate and hyper realistic details. Beautifully color graded with modern and cinematic light. Captured with a Phase One XF IQ4 camera, 200 Mega Pixels, it features insane detailing and depth of field. The textures give a feeling of depth and richness, enhancing the overall beauty of the composition. The editorial photography and photoshoot elements are evident in the detailed and professional capture. \\
\hline
SC-general, Row 1, Col 3 & White and grayish Persian cat with fluffy fur, vibrant green eyes, not a flat nose, has a distinct stop, looking directly into the camera, soft dramatic lighting, cinematic style, slightly backlit. \\
\hline
SC-general, Row 1, Col 4 & an alien cyborg with eyes and oozing in the woods, in the style of rendered in cinema4d, undefined anatomy, tangled nests, dark white and crimson, eerily realistic, soft sculptures, made of mist. \\
\hline
SC-general, Row 1, Col 5 & A luminous figure draped in glowing robes holds a radiant orb of light with plants and leaves on their shoulders, resembling the Keeper of the Light, Dota 2, in an enchanting, mystical forest ambiance. \\
\hline
SC-artext, Row 1, Col 1 & The image presents a simple yet striking visual. Dominating the frame is the word ``Average", spelled out in capital letters. Each letter is identical in size and color, creating a sense of uniformity and balance. The letters are not solid but rather composed of small bumps, giving them a textured appearance that stands out against the stark white background. The word ``Average" is centrally positioned, drawing the viewer's attention immediately to it. Despite the simplicity of the elements involved, the image conveys a clear message: the word ``Average". The absence of any other elements or distractions underscores this message, making it the sole focus of the viewer's attention. \\
\hline
SC-artext, Row 1, Col 2 & The image presents a 3D rendering of the word ``Average". The word is written in a cursive font and is colored in a vibrant shade of blue. It's slightly tilted to the right, adding a dynamic touch to the overall composition. Each letter is slightly larger than the last, creating a cascading effect that leads the viewer's eye down the word. The background is a stark white, which contrasts sharply with the blue of the word, making it stand out prominently. The image does not contain any other objects or text, and the focus is solely on the word "Average". The simplicity of the image allows the viewer to clearly see and understand the meaning of the word. \\
\hline
SC-artext, Row 1, Col 3 & The image presents a 3D rendering of the word ``Average". The word is written in a bold, sans-serif font and is colored in a vibrant shade of red. The letters are slightly tilted to the right, adding a dynamic touch to the overall composition. Each letter is enveloped in a ring of fire, with the letters ``A", ``V", and ``R" being particularly noticeable due to their larger size. The background is a stark white, which contrasts sharply with the fiery red of the word, making it stand out prominently. The image does not contain any other discernible objects or text. The focus is solely on the word ``Average" and its fiery presentation. \\
\hline
SC-artext, Row 1, Col 4 & The image presents a 3D rendering of the word ``Average" in a vibrant shade of green. The letters are intricately crafted from grass, giving them a natural and organic feel. Each letter is adorned with small white flowers, adding a touch of whimsy to the overall design. The letters are arranged in a staggered formation, creating a sense of depth and dimension. The word ``Average" stands out prominently against the stark white background, making it the focal point of the image. The image does not contain any discernible text apart from the word ``Average". \\
\hline
SC-artext, Row 1, Col 5 & The image presents a vibrant display of the word ``Average" in a cursive font. The letters are filled with splashes of paint in a rainbow of colors, transitioning from red to orange, then to yellow, green, blue, and finally to purple. Each letter is slightly tilted, adding a dynamic feel to the overall composition. The background is a stark white, which contrasts with the colorful text and allows it to stand out prominently. The word ``Average" is the only text present in the image. The relative positions of the letters suggest they are stacked on top of each other, further enhancing the visual impact of the image. \\
\bottomrule
\end{tabularx}
\vspace{-2mm}
\caption{Examples of texts in SC-general and SC-artext. Textual description of the first row in Figure \ref{fig:supp-3}.}
\label{tab:prompt-list}
\vspace{-6mm}
\end{table*}

\end{document}